\documentclass[10pt,twocolumn,letterpaper]{article}
\usepackage{iccv}
\usepackage{times}
\usepackage{epsfig}
\usepackage{graphicx}
\usepackage{amsmath}
\usepackage{amssymb}
\usepackage[dvipsnames]{xcolor}
\usepackage{booktabs}
\usepackage{overpic}
\usepackage{float}
\usepackage[linesnumbered, boxed, ruled]{algorithm2e}
\usepackage{multirow}
\usepackage{array}
\usepackage{soul}
\usepackage{dirtytalk}
\usepackage{overpic}
\usepackage{mathtools}
\usepackage{subfig}
\usepackage[affil-it]{authblk}

\usepackage[pagebackref=true,breaklinks=true,letterpaper=true,colorlinks,bookmarks=false]{hyperref}

\iccvfinalcopy 

\def\iccvPaperID{6267} 

\ificcvfinal\fi

\newif\ifdraft
\draftfalse

\ifdraft
\definecolor{darkg}{rgb}{0,0.4,0}
\newcommand{\dcc}[1]{{\color{red}[\textbf{Danny:} #1]}}
\newcommand{\ahc}[1]{{\color{purple}[\textbf{Amir:} #1]}}
\newcommand{\kac}[1]{{\color{blue}[\textbf{Kfir:} #1]}}

\newcommand{\ah}[1]{{\color{purple}#1}}
\newcommand{\ka}[1]{{\color{blue}#1}}

\newcommand{\drop}[1]{}

\else
\newcommand{\ahc}[1]{}
\newcommand{\dcc}[1]{}
\newcommand{\kac}[1]{}
\newcommand{\ypc}[1]{}

\newcommand{\ah}[1]{{\color{black}#1}}
\newcommand{\ka}[1]{{\color{black}#1}}

\fi

\newcommand{\ourmethod}{Delta Denoising Score}
\newcommand{\om}{DDS}

\makeatletter
\DeclareRobustCommand\onedot{\futurelet\@let@token\@onedot}
\def\@onedot{\ifx\@let@token.\else.\null\fi\xspace}

\newcommand{\textcond}{y}
\newcommand{\textref}{\hat{\textcond}}
\newcommand{\latent}{\mathbf{z}}
\newcommand{\noised}{\mathbf{z_t}}

\newcommand{\latentref}{\mathbf{\hat{z}}}
\newcommand{\noisedref}{\mathbf{\hat{z}_{t}}}

\newcommand{\cfg}{\omega}
\newcommand{\task}{k}

\begin{document}

\title{\ourmethod{}} 


\author[1,2]{\hspace{30pt}Amir Hertz\footnote{}}
\author[1]{\hspace{30pt}Kfir Aberman}
\author[1,2]{\hspace{20pt} Daniel Cohen-Or\textsuperscript{*}}
\affil[1]{Google Research} 
\affil[2]{Tel Aviv University}

\maketitle
\ificcvfinal\thispagestyle{empty}\fi


\global\csname @topnum\endcsname 0
\begin{figure}[t]
\footnotesize
\centering
    \begin{overpic}[width=1\columnwidth,tics=10, trim=0mm 0 0mm 0,clip]{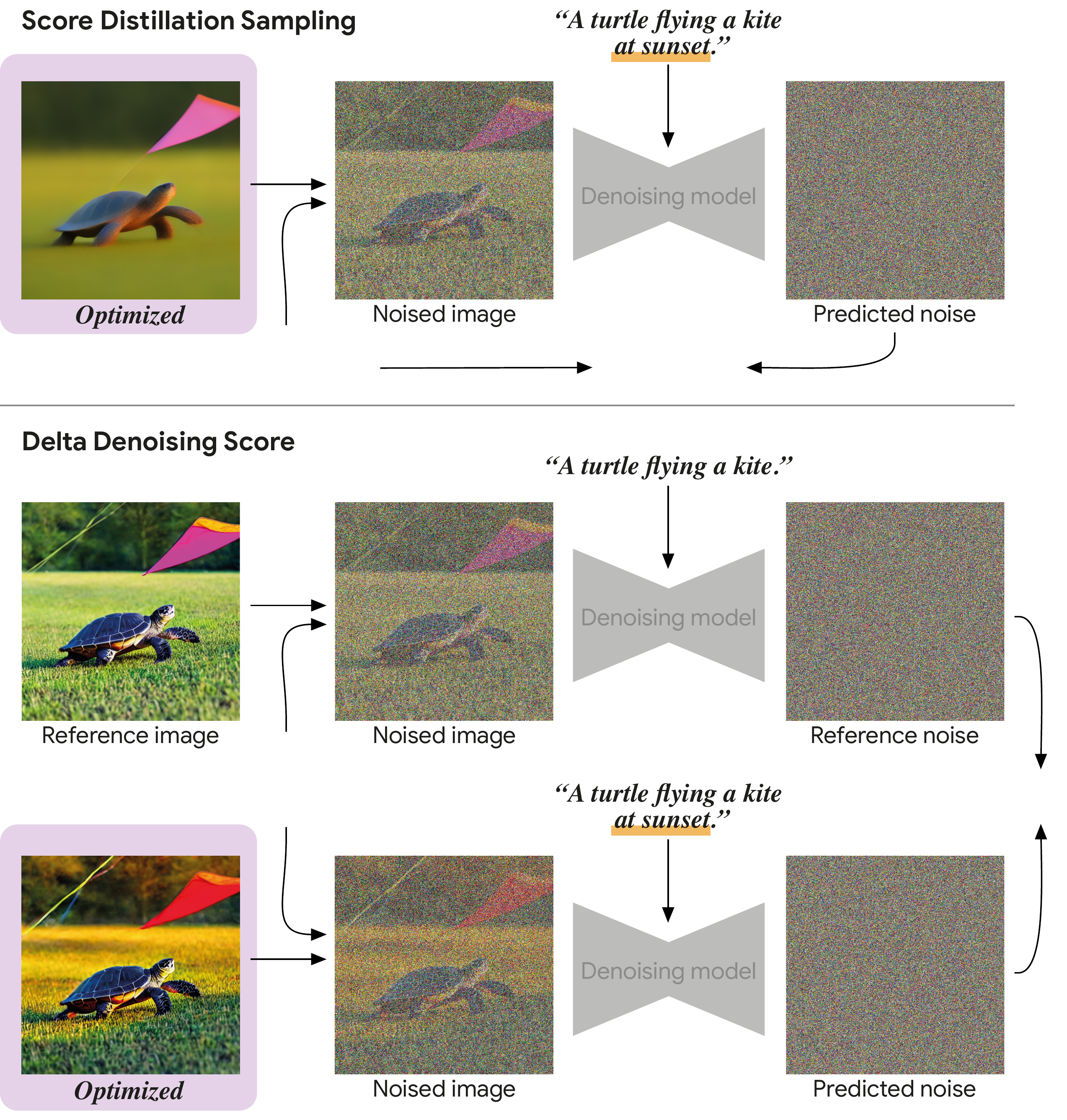}
    \put(10.5,55.5){$\latentref$}
    \put(10.5,24){$\latent$}
     \put(10.5,93){$\latent$}
     
     \put(38,55.8){$\noisedref$}
    \put(38,24.3){$\noised$}
     \put(38,93.3){$\noised$}

      \put(78.2,56.3){$\hat{\epsilon}_{\phi}$}
      \put(78.2,24.9){$\epsilon_{\phi}$}
      \put(78.2,93.8){$\epsilon_{\phi}$}
     
     \put(20.5,29.5){$\substack{t \sim \mathcal{U}(0, 1) \\ \epsilon \sim \mathcal{N}(0, \mathbf{I})}$}
     \put(20.5,66.){$\substack{t \sim \mathcal{U}(0, 1) \\ \epsilon \sim \mathcal{N}(0, \mathbf{I})}$}

      \put(87,28){$\nabla_\theta \mathcal{L}_\textrm{DDS}$}
      \put(54,66.1){$\nabla_\theta \mathcal{L}_\textrm{SDS}$}
    \end{overpic}
\vspace{2pt}
\caption{\bf{Score Distillation Sampling (SDS) vs. \ourmethod~(\om).} {\it 
Top: SDS mechanism optimizes a given image by querying the denoising model on the noisy version of the image and a target text prompt. The resulting image can often be blurry and unfaithful to the target prompt.
Bottom: DDS queries an additional reference branch with a matched text-prompt, and generates delta scores that represent the difference between the outputs of the two queries. DDS provides cleaner gradient directions that modify the edited portions of the optimized image, while leaving the other parts unchanged.}}
\label{fig:teaser}
\end{figure}
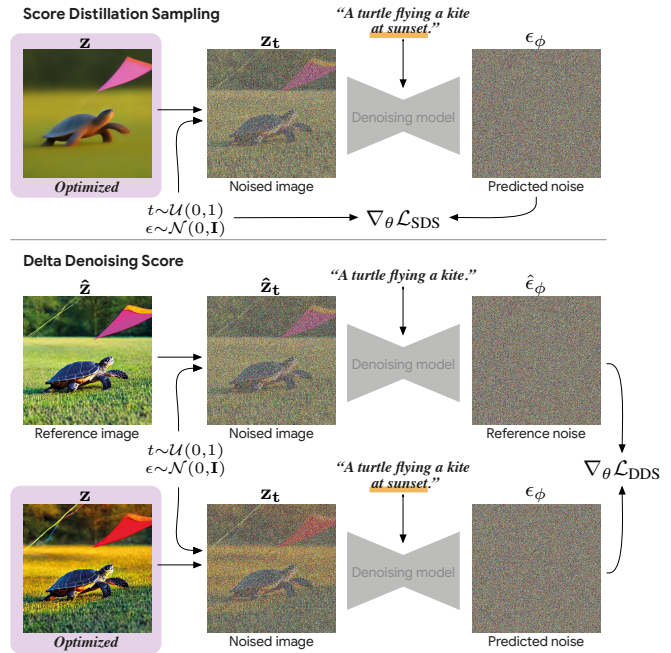


     

     


\begin{abstract}
We introduce Delta Denoising Score (DDS), a novel scoring function for text-based image editing that guides minimal modifications of an input image towards the content described in a target prompt. DDS leverages the rich generative prior of text-to-image diffusion models and can be used as a loss term in an optimization problem to steer an image towards a desired direction dictated by a text.
DDS utilizes the Score Distillation Sampling (SDS) mechanism for the purpose of image editing. We show that using only SDS often produces non-detailed and blurry outputs due to noisy gradients. To address this issue, DDS uses a prompt that matches the input image to identify and remove undesired erroneous directions of SDS. Our key premise is that SDS should be zero when calculated on pairs of matched prompts and images, meaning that if the score is non-zero, its gradients can be attributed to the erroneous component of SDS.
Our analysis demonstrates the competence of DDS for text based image-to-image translation. We further show that DDS can be used to train an effective zero-shot image translation model. Experimental results indicate that DDS outperforms existing methods in terms of stability and quality, highlighting its potential for real-world applications in text-based image editing. 
For code and additional results, please visit our project page: \url{https://delta-denoising-score.github.io/}.
\end{abstract}

\footnotetext[1]{Performed this work while working at Google.\label{note1}}

\section{Introduction}

Large-scale language-vision models have revolutionized the way images and visual content, in general, can be generated and edited. Recently, we have witnessed a surge in the development of text-to-image generative models, which utilize textual input to condition the generation of images.
A promising avenue in this field is Score Distillation Sampling (SDS) \cite{poole2022dreamfusion} -- a sampling mechanism that utilizes probability density distillation to optimize a parametric image generator using a 2D diffusion model as a prior. 

The effectiveness of the SDS stems from rich generative prior of the diffusion model it samples from. 
This is in contrast to the direct use of a language-vision model, like CLIP, which was trained using contrastive loss \cite{radford2021learning}.
The prior of large generative diffusion models, like Stable Diffusion~\cite{rombach2021highresolution}, DALLE-2~\cite{ramesh2022hierarchical} and Imagen~\cite{saharia2021image} is particularly rich and expressive and has been demonstrated to be highly effective in generating visually stunning assets across various domains, including images and 3D models, among others.

Despite its usefulness, one of the primary issues associated with SDS is its tendency to converge towards specific modes, which often leads to the production of blurry outputs that only capture the elements explicitly described in the prompt. 

In particular, using SDS to \emph{edit an existing image} by initializing the optimization procedure from that image, may result in significant blurring of the image beyond the edited elements.

In this paper, we introduce a new diffusion-based scoring technique for optimizing a parametric model for the task of editing. 
Unlike SDS, which queries the generative model with a pair of image and text, \ka{our method utilizes an additional query of a reference image-text pair, where the text matches the content of the image}.  Then, the output score is the difference, or \textit{delta}, between the results of the two queries (see Figure~\ref{fig:teaser}). We refer to this scoring technique as \ourmethod{} (\om). 

In its basic form, \om{} is applied on two pairs of images and texts, one is a reference image-text that remains intact during the optimization, and the other is a target image that is optimized to match a target text prompt. The delta scoring provides effective gradients, which modify the edited portions of the image, while leaving the others unchanged. 

The key idea is that the source image and its text description, can be used for estimating  undesirable and noisy gradients directions introduced by SDS. Then if we want to alter only a portion of the image using a new text description, we can use our reference estimation and get a cleaner gradient direction to update the image.

\om{} can be used as a prompt-to-prompt editing technique that can modify images by only editing their captions, where no mask is provided or computed. Beyond that, \ourmethod{} enables us to train a distilled image-to-image model without the need of paired training dataset, yielding a zero-shot image translation technique.
Training the model, requires only dataset of the source distribution, associated with simple captions that describe the source and target image distributions.  As we will show, such zero-shot training can be applied on a single or multi-task image translation, and the source distribution can include synthetically generated and real images.

To demonstrate the effectiveness of our approach, we conducted experiments comparing our model to existing state-of-the-art text-driven editing techniques.

\begin{figure}
    \includegraphics[width=\columnwidth]{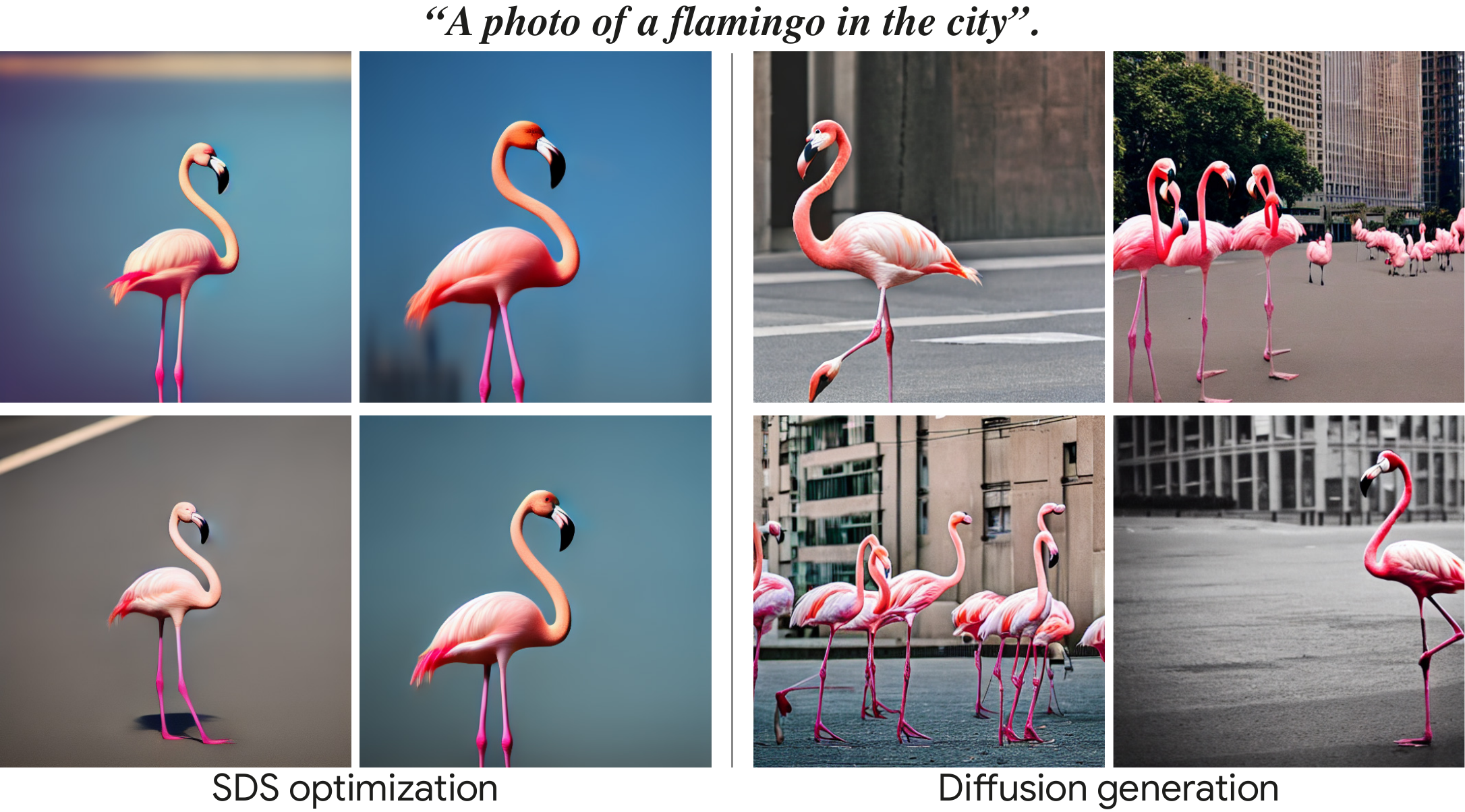}
    \caption{\bf{Sampling text-to-image diffusion models.} {\it Generation via SDS optimization  starting from random noises (left) vs. conventional diffusion-based image generation (right). Both samples are generated with respect to a given text prompt (top). Generating images based on SDS only leads to less diverse results and mode collapse where the main subject in the text appears in front of a blurry background.}}
    \label{fig:flamingo}
\end{figure}

\section{Related Work}


Text-to-Image models~\cite{saharia2022photorealistic,ramesh2022hierarchical,rombach2021highresolution}, have recently raised the bar for the task of generating images conditioned on a text prompt, exploiting the powerful architecture of diffusion models \cite{ho2020denoising,sohl2015deep,song2019generative,ho2020denoising,song2020denoising,rombach2021highresolution}, which can be used to various image editing and guided synthesis tasks~\cite{ruiz2022dreambooth,Kawar2022ImagicTR,voynov2023p+,voynov2022sketch}.

Recent works have attempted to adapt text-guided diffusion models to the fundamental challenge of single-image editing, aiming to exploit their rich and diverse semantic knowledge.
Meng et al.~\cite{meng2021sdedit} add noise to the input image and then perform a text-guided denoising process from a predefined step. Yet, they struggle to accurately preserve the input image details, which were preserved by a user provided mask in other works~\cite{nichol2021glide, avrahami2022blended, avrahami2022blendedlatent}. DiffEdit \cite{couairon2022diffedit} uses DDIM inversion for image editing, but avoids the emerged distortion by automatically producing a mask that allows background preservation.

While some text-only editing approaches are bound to global editing  \cite{crowson2022vqgan, kwon2021clipstyler, kim2022diffusionclip,patashnik2021styleclip}, Bar-Tal et al.~\cite{bar2022text2live} propose a text-based localized editing technique without using any mask. Their technique allows high-quality texture editing, but not modifying complex structures, since only CLIP \cite{radford2021learning} is employed as guidance instead of a generative diffusion model. Prompt-to-prompt~\cite{hertz2022prompt} suggests an intuitive editing technique that enables manipulation of local or global details for images that were synthesized by a text-to-image network. \cite{mokady2022null} proposed an approach to invert real images into the latent space of the diffusion model, such that prompt-to-prompt can be applied to real images.
Imagic \cite{Kawar2022ImagicTR} and UniTune\cite{valevski2022unitune} have demonstrated impressive text-driven editing capabilities, but require the costly fine-tuning of the model. InstructPix2Pix~\cite{brooks2022instructpix2pix}, plug-and-play~\cite{tumanyan2022plug} and \cite{parmar2023zero} can get an instruction or target prompt and manipulate real images towrds the desired edit

DreamFusion~\cite{poole2022dreamfusion} proposed the SDS score as a 2D prior which can be used to generate 3D assets~\cite{metzer2022latent,raj2023dreambooth3d}. SDS is also used in~\cite{song2022diffusion} to direct a StyleGAN generator for the domain adaption task. This is conceptually similar to StyleGAN-NADA~\cite{gal2021stylegan} which uses instead CLIP~\cite{radford2021learning} to translate the domain of a StyleGAN generator to other domains based only textual description. Our work explores the usage of SDS score in the context of image editing, and propose a new technique to clean the undesired gradients of SDS which grab the optimization process  into noisy direction that smooth out relevant detailed from the original image.

\begin{figure}
    \includegraphics[width=\columnwidth]{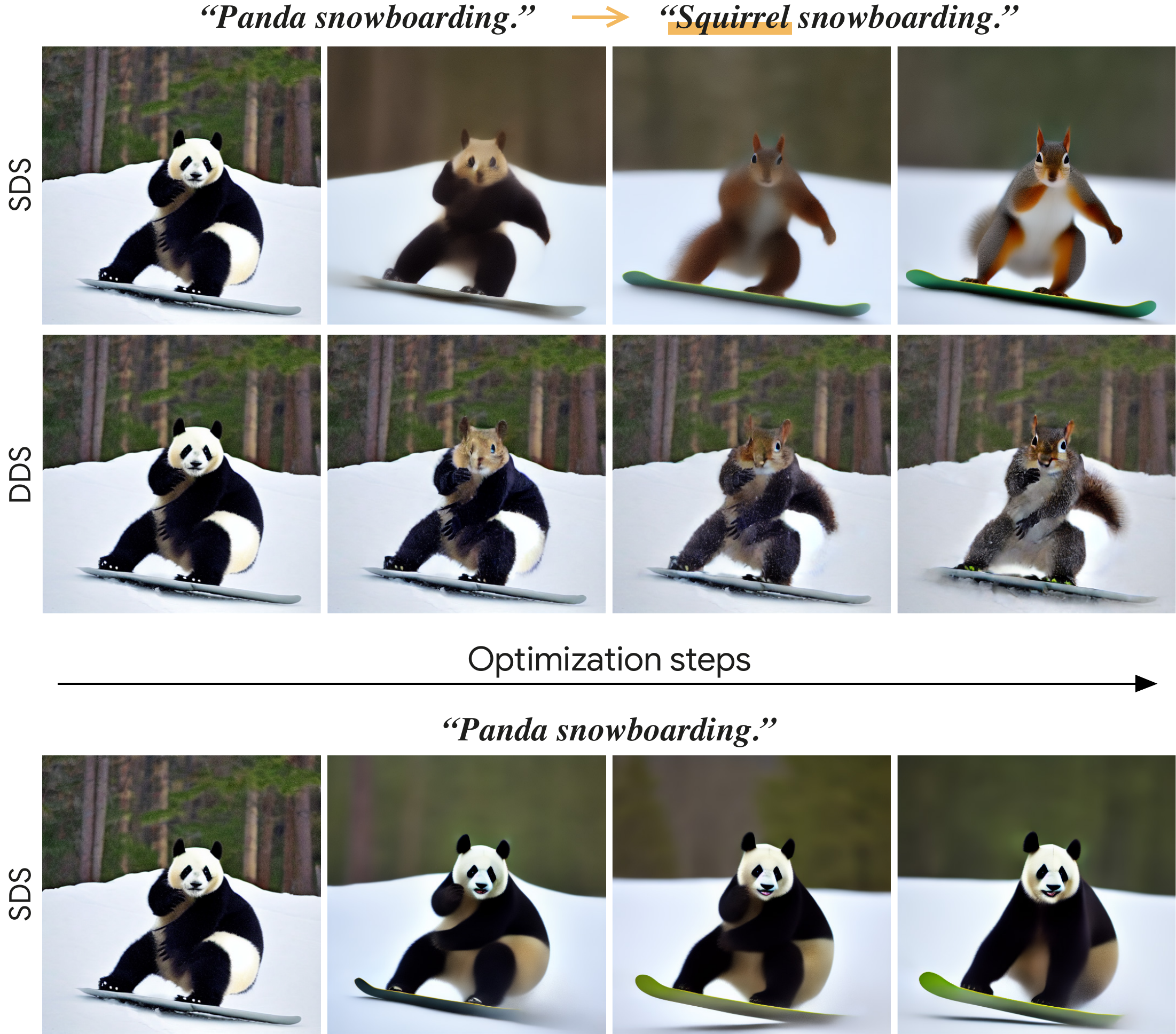}
    \caption{\bf{Bias in SDS optimization}. {\it Left column: an image generated by the prompt ``Panda snowboarding". Top rows show the difference between SDS to DDS optimization when changing the animal in the prompt (``Panda" to ``Squirrel").
     Bottom row shows SDS optimization applied using the original prompt. Even in this case, the image becomes blurry.}}
    \label{fig:panda_snow}
\end{figure}

\section{Delta Denoising Score (DDS)}
\label{section:method}

We begin with a brief overview of the SDS loss function and explain the challenges in sampling and editing images with SDS, based on empirical observations. In particular, we demonstrate that SDS introduces a noisy direction when applied to the task of image editing. We then  introduce our \ourmethod~(\om), which utilizes a reference pair of image and text to correct the noisy direction of SDS and offers a new technique for the task of prompt-to-prompt editing \cite{hertz2022prompt}. 
We conduct all our experiments using the latent model-- Stable Diffusion \cite{rombach2021highresolution}, nevertheless, in our overview and results, we refer to the models latents and output channels as images and pixels respectively.

\subsection{SDS overview}
Given an input image $\latent$, a conditioning text embedding $y$, a denoising model $\epsilon_{\phi}$ with parameters set $\phi$, a randomly sampled timestep $t\sim\mathcal{U}(0, 1)$ drawn from the uniform distribution, and noise $\epsilon\sim\mathcal{N}(0, \mathbf{I})$ following a normal distribution, the diffusion loss can be expressed as:
$$
\mathcal{L}_{\text{Diff}}\left(\phi, \latent, \textcond, \epsilon, t  \right) = w(t)||\epsilon_{\phi}\left(\noised, \textcond, t\right) - \epsilon||_{2}^{2},
$$
where $w(t)$ is a weighting function, and $\noised$ refers to the noisy version of $\latent$ obtained via a stochastic noising forward process given by $\noised = \sqrt{\alpha_t} \latent + \sqrt{1-\alpha_t} \epsilon$, with $\alpha_t$ being the noise scheduler. For simplicity, we omit the weighting factor in the remainder of this section.

The text conditioned diffusion models use classifier-free guidance (CFG \cite{ho2022classifier}) that consists of two components, one that is conditioned on text input, and another that is unconditioned. During inference, the two components are used to denoise the image via
$$
\epsilon_{\phi}^{\cfg}\left(\noised, \textcond, t\right) = \left(1 + \cfg\right) \epsilon_{\phi}\left(\noised, \textcond, t\right) - \cfg \epsilon_{\phi}\left(\noised, t\right),
$$
where the components are balanced using a guidance parameter $\omega$.

Given an arbitrary differentiable parametric function that renders images, $g_{\theta}$, 
the gradient of the diffusion loss function with respect to the parameters $\theta$ is given by:

$$
\nabla_\theta \mathcal{L}_{\text{Diff}} = \left(\epsilon_{\phi}^{\cfg}\left(\noised, \textcond, t\right) - \epsilon  \right)  \dfrac{\partial{\epsilon_{\phi}^{\cfg}\left(\latent, \textcond, t\right)}}{\partial{\noised}}\dfrac{\partial\noised}{\partial\theta}.
$$ 

It has been demonstrated in~\cite{poole2022dreamfusion} that omitting the U-Net Jacobian term (middle term) leads to an effective gradient for optimizing a parametric generator with diffusion models:

\begin{equation}
\nabla_\theta \mathcal{L}_{\text{SDS}}(\latent, y, \epsilon, t)= \epsilon_{\phi}^{\cfg}\left(\left(\noised, \textcond, t\right) - \epsilon  \right) \dfrac{\partial\noised}{\partial\theta}.
\end{equation}

Incrementally updating the  parameters of the generator in the direction of the gradient, produces images that exhibit a higher degree of fidelity to the prompt. However, SDS suffers from the tendency to converge towards specific modes any relevant citation in mind, resulting in non-diverse and blurry outputs that only highlight elements mentioned in the prompt. Figure~\ref{fig:flamingo} showcases a comparison between sampling Stable-Diffusion with SDS vs. sampling it with a standard reverse process of the diffusion model, demonstrating this issue with 2D image samples.

The original purpose of SDS was to generate samples via optimization from a text-conditioned diffusion model. It is noteworthy that $g_{\theta}$ can be an arbitrary parametric function that renders images. In the following sections we demonstrate our results with  $g_{\theta} = \theta$, namely, a trivial generator that renders a single image, where the optimization variables are the image pixels themselves, however, note that the derivation is general.

\begin{figure}
    \footnotesize
   \centering\begin{overpic}[width=\columnwidth,tics=10, trim=0mm 0 0mm 0,clip]{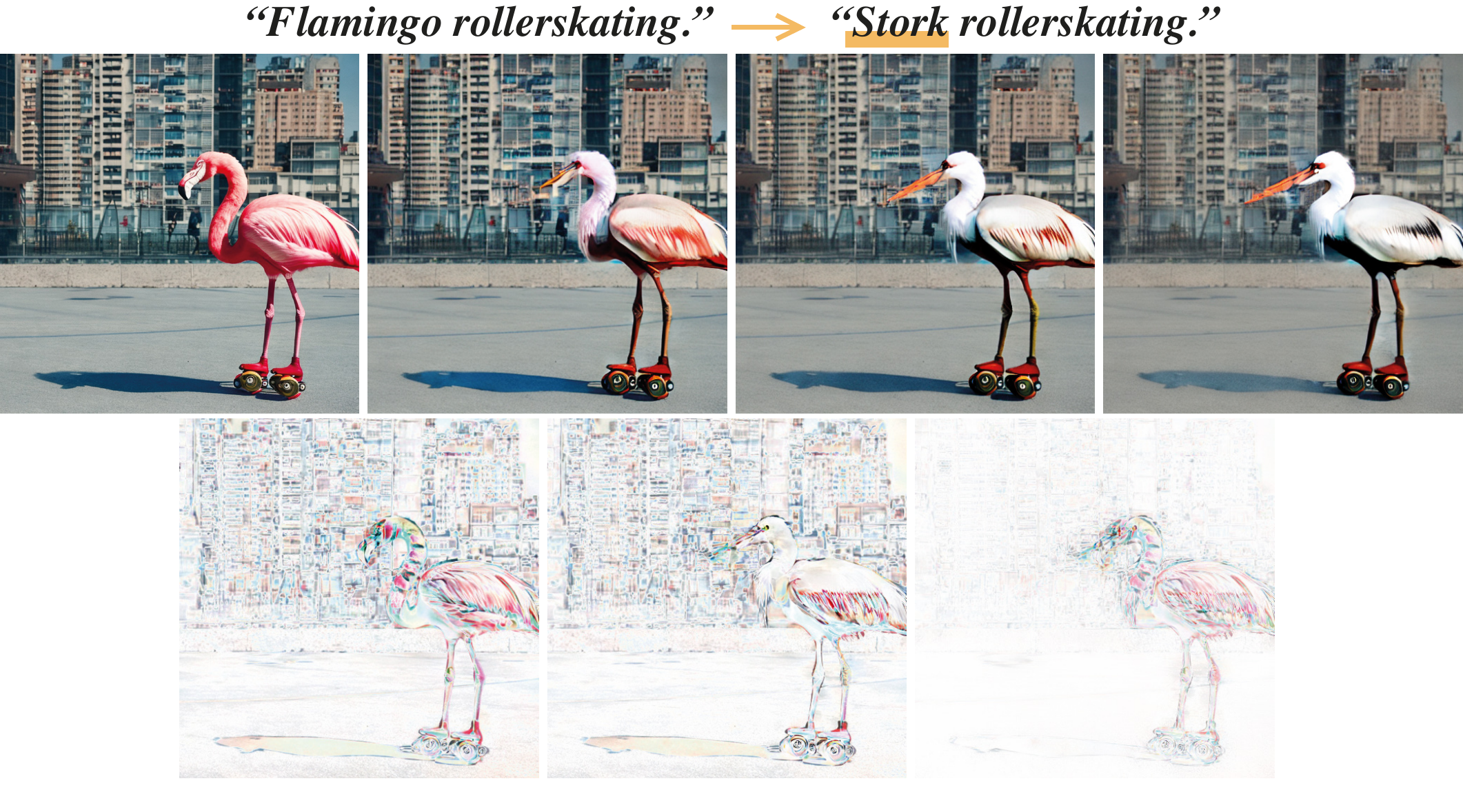}
   \put(15,-.5){$\nabla_{\theta} \mathcal{L}_{\text{SDS}} (\hat{\latent}, \hat{y})$}
   \put(40,-.5){$\nabla_{\theta} \mathcal{L}_{\text{SDS}} (\latent, y)$}
    \put(69,-.5){$\nabla_{\theta} \mathcal{L}_{\text{DDS}}$}
   \end{overpic}
    \caption{\bf{DDS gradients.} \it{Top: Visualization of 4 steps in the DDS optimization process, where an image of a ``Flamingo rollerskating" (left) gradually transforms into a ``Stork rollerskating" (right). Bottom: By subtracting the SDS gradients of the reference image and the source prompt (left) from the SDS gradients of the edited image and the target prompt (middle), we obtain cleaner DDS gradients (right).}
    }
    \label{fig:grad_vis}
\end{figure}
\subsection{Editing with SDS}

The original purpose of SDS was to generate samples from a distribution conditioned solely on a text prompt. However, we now aim to extend SDS to the task of editing, which involves conditioning the sampling process on both an image and text.

Our objective is to synthesize an output image $\latent$ that incorporates the structure and details of an input source image $\latentref$, while conforming to the content specified in a target prompt $y$. This is a standard text-driven image-to-image translation problem, where modifications may be applied locally or globally \cite{hertz2022prompt,brooks2022instructpix2pix}.

One potential approach to utilize SDS is to initialize the optimization variable with the source image $\latent_0=\latentref$ and applying SDS while conditioning on $y$. However, we have observed that similarly to the non image conditioned SDS, this approach leads to blurred outputs and a loss of details, particularly those that are unrelated to the input prompt. Figure~\ref{fig:panda_snow} (top row) demonstrates such example where the panda transforms into a squirrel at the cost of blurring out other details.

Based on our observations, we define a decomposition for the gradients $\nabla_{\theta} \mathcal{L}_{\text{SDS}}$ to two components:
one component $\delta_\textrm{text}$ is a desired direction that directs the image to the closest image that \ka{matches}  the text. And another, undesired component, $\delta_\textrm{bias}$ that interferes with the process and causes the image to become smooth and blurry in some parts. Formally:
\begin{equation}
\nabla_{\theta} \mathcal{L}_{\text{SDS}} (\latent, y, \epsilon, t) \coloneqq  \delta_\textrm{text} + \delta_\textrm{bias},
\label{textandbias}
\end{equation}
where both $\delta_\textrm{text}$ and $\delta_\textrm{bias}$ are random variables that depend on $\latent$, $\textcond$,  $\epsilon$ and $t$.
Under this definition, to address this issue and enable high-quality or distilled image editing with SDS, we have to isolate and extract the text-aligned part  $\delta_\textrm{text}$ and follow it during the optimization while avoiding the $\delta_\textrm{bias}$ direction that may take the image to unintended places.

\subsection{Denoising the Editing Direction}

We next aim to find the noisy direction of the SDS score, when applied for editing purposes, and remove it during the optimization process.

The gist of our method is that since we already have a source image and its text description, 
they can be used for estimating the noisy direction $\delta_\textrm{bias}$, that biases the edit towards undesired directions.
Then, if we want to alter only a portion of the image using a new text description, we can use our reference estimation and get a \textit{cleaner} gradient direction to update the image.
In practice, we use a reference branch that calculates the SDS score of the given image  $\latentref$ with a corresponding, matched, text prompt $\hat{y}$, and subtract it from the main SDS optimization branch to yield a distilled edit.

Formally, given \ka{matched and unmatched} image-text embedding pairs $\latentref$, $\textref$,  $\latent$, $\textcond$  respectively, the delta denoising loss is given by:
$$
\mathcal{L}_{\text{DD}}\left(\phi, \latent, \textcond, \latentref, \textref, \epsilon, t  \right) =
||\epsilon_{\phi}^{\cfg}\left(\noised, \textcond, t\right) - \epsilon_{\phi}^{\cfg}\left(\noisedref, \textref, t\right)||_{2}^{2},
$$
where $\noised$ and $\noisedref$ share the same sampled noise $\epsilon$ and timestep $t$.
Then, the gradient over $g_\theta = \latent$, are given by
$$\nabla_{\theta} \mathcal{L}_{\text{DD}} = \left( \epsilon_{\phi}^{\cfg}\left(\noised, \textcond, t\right) - \epsilon_{\phi}^{\cfg} \left(\noisedref, \textref, t\right)\right)  \dfrac{\partial{\epsilon_{\phi}^{\cfg}\left(\noised, \textcond, t\right)} }{\partial{\noised}}\dfrac{\partial\latent}{\partial\theta}.$$ 

Again, we omit the differentiation thorough the diffusion model to obtain the Delta Denoising Score,
\begin{equation}
\nabla_{\theta} \mathcal{L}_{\text{DDS}} = \left( \epsilon_{\phi}^{\cfg}\left(\noised, \textcond, t\right) - \epsilon_{\phi}^{\cfg} \left(\noisedref, \textref, t\right)\right)  \dfrac{\partial\latent}{\partial\theta}.
\label{Eq:sds_dir}
\end{equation}

We state that DDS pushes the optimized image into the direction of the target prompt without the interference of the noise component, namely, $\nabla_{\theta} \mathcal{L}_{\text{DDS}} \approx \delta_{\text{text}}$. 

By adding and subtracting $\epsilon$ from the term in~\eqref{Eq:sds_dir}, we can represent DDS as a difference between two SDS scores:

\begin{eqnarray}
\nabla_{\theta} \mathcal{L}_{\text{DDS}} & = &  \nabla_{\theta} \mathcal{L}_{\text{SDS}} (\latent, y)-   \nabla_{\theta} \mathcal{L}_{\text{SDS}} (\hat{\latent}, \hat{y}).
\label{eq:dds}
\end{eqnarray}

We first claim that the score provided by the reference branch is equivalent to the noisy direction. This is because, ideally, a matched image-text pair should have a low average SDS gradient across various timesteps and noise instances. Therefore, any non-zero gradient can be attributed to the noisy direction, thus, 
\begin{equation}
    \nabla_\theta \mathcal{L}_{\text{SDS}}(\latentref,\hat{y}) = \hat{\delta}_{\text{bias}} .
\label{eq:sds_is_noise}
\end{equation}

Evidently, the score of a \ka{matched} text-to-image pair is non-zero. As can be seen in Figure~\ref{fig:panda_snow} (bottom, row), even when the optimization process starts with an image that was generated by the text, there are gradients that pull the image towards the non-desired modes. For further empirical results of the estimation of $\delta_{\text{bias}}$, please refer to Section~\ref{sec:results_sds}.

\begin{figure}
    \scriptsize
   \centering\begin{overpic}[width=.49\columnwidth,tics=10, trim=0mm 0 0mm 0,clip]{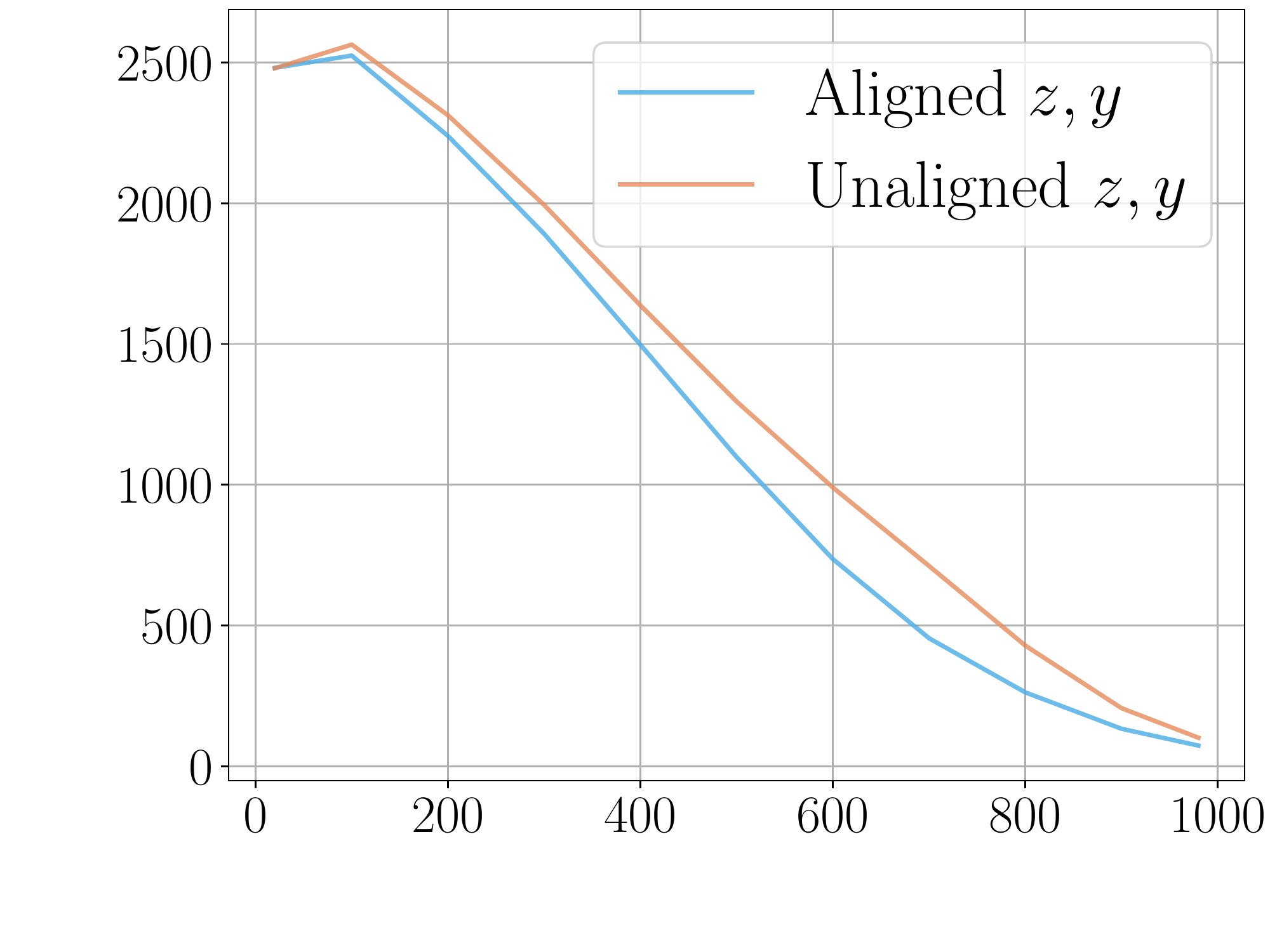}
    \put(43,1){Timestep $t$}
    \put(1,29){\rotatebox{90}{SDS Norm}}
    \end{overpic}
    \begin{overpic}[width=.49\columnwidth,tics=10, trim=0mm 0 0mm 0,clip]{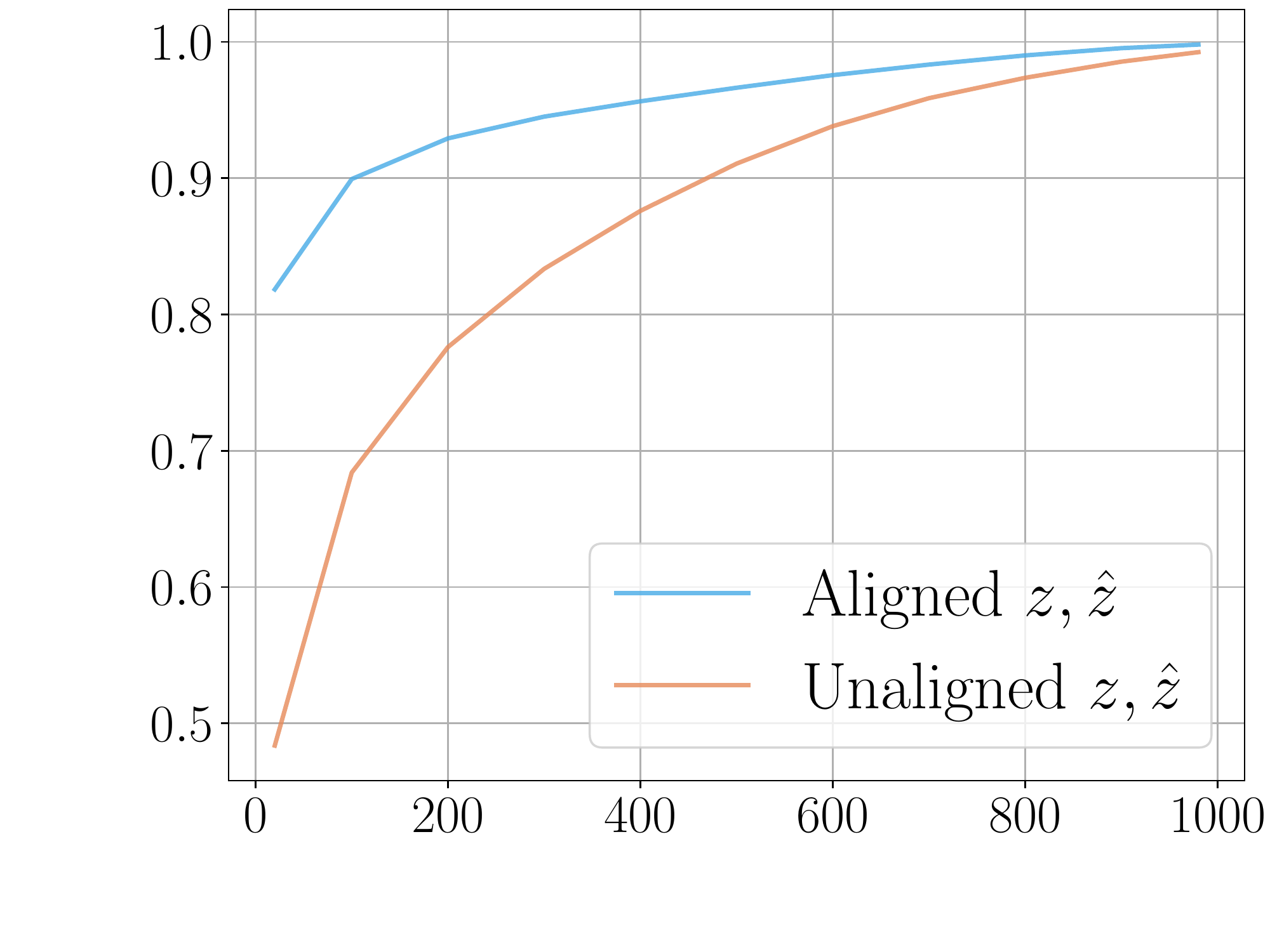}
    \put(43,1){Timestep $t$}
    \put(3.5,29){\rotatebox{90}{DDS Cosine}}
    \end{overpic}
    \caption{{\bf Expected SDS gradients.} \it{Left: Expected SDS norm $||\nabla \mathcal{L}_{\text{SDS}}(z, y)||_2$ across different timesteps for matched (blue curve) and unmatched (orange curve) synthetic image-text pairs. 
    Right: Cosine similarity between the SDS directions in \eqref{eq:dds} on matched (blue) and unmatched (orange) images from the InstructPix2Pix dataset~\cite{brooks2022instructpix2pix}.}
    }
    \vspace{-5pt}
    \label{fig:sds_graph}
\end{figure}

We next claim that the noisy component $\delta_{\text{noise}}$ of closely related images (e.g., images with similar structure that were created with close prompts) is similar. This is demonstrated in the DDS evaluation experiment in Section~\ref{sec:exp} and in Figure~\ref{fig:sds_graph} which shows that the consine similarity between the directions of the matched pair is high.  
This means that $\delta_\textrm{bias} \approx \hat{\delta}_\textrm{bias}$.

By combining the conclusions drawn from the above-mentioned experiments, we get $\nabla_{\theta} \mathcal{L}_{\text{DDS}} \approx  \hat{\delta}_\textrm{text}$, which indicates that our DDS can be considered a distilled direction that concentrates on editing the relevant portion of the image, such that it \ka{matches} the target text.

Figure~\ref{fig:grad_vis} visualizes the key idea behind DDS, The figure shows the two noisy SDS scores, of the \ka{matched and unmatched} pair, along with their difference, which comprises DDS. Notably, subtracting the two noisy scores produces a clear and concise score that concentrates solely on the targeted modification in the image.
\begin{figure}[t]
\footnotesize
\centering
    \begin{overpic}[width=1\columnwidth,tics=10, trim=0mm 0 0mm 0,clip]{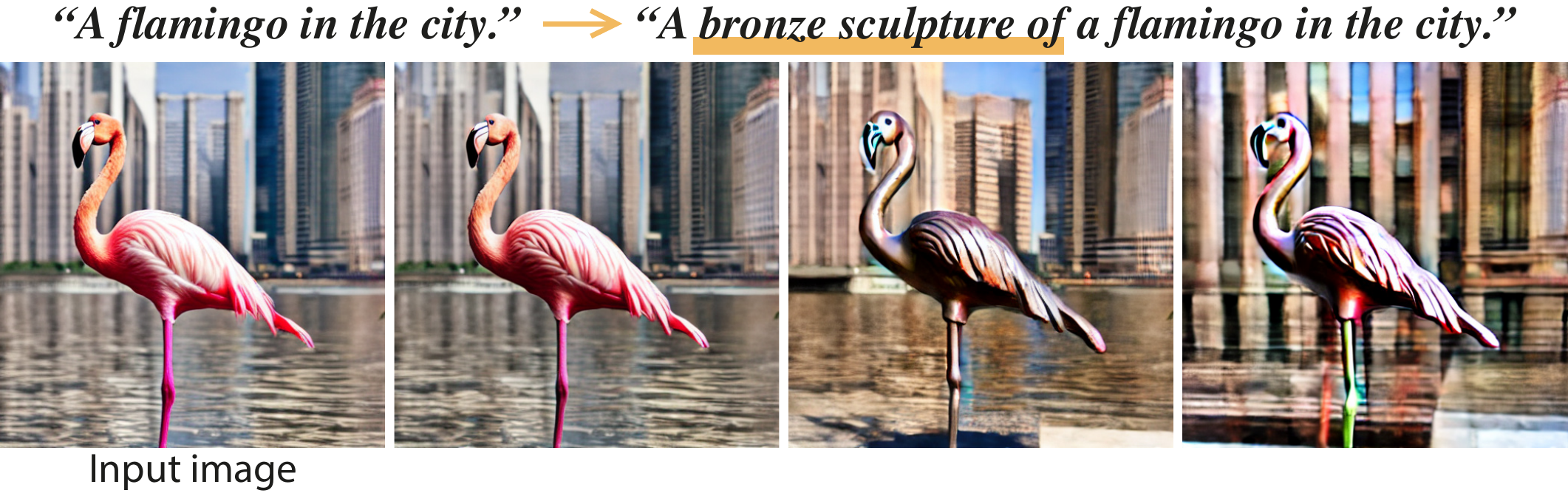}
   
      \put(34,0){$\cfg{}=3$}
      \put(61,0){$15$}
      \put(86,0){$30$}
    \end{overpic}
\caption{{\bf \om{} optimization results using different values of the classifier-free guidance scale $\cfg{}$.} {\it On one hand, using small values of $\cfg{}$ leads to slow convergence and low fidelity to the text prompt. On the other hand, using large values of $\cfg{}$ results in low fidelity to the input image.}}
\vspace{-10pt}
\label{fig:flamingo_cfg}
\end{figure}

\paragraph{Effect of CFG on DDS}
As previously noted, the Classifier Free Guidance (CFG) parameter $\cfg{}$, regulates the relative influence of the text-conditioned and unconditional components of the denoising objective. Apparently, despite the subtraction of the two distinct branches in~\om{}, $\cfg{}$ still has a discernible impact on the resulting image output. Our experiments show that small values of $\cfg{}$ yield slower convergence rates and a correspondingly diminished fidelity to the text prompt, while larger $\cfg{}$ values result in an attenuated fidelity to the input image. This observed phenomenon is visualized in Figure~\ref{fig:flamingo_cfg}
\ah{and empirically evaluated in Section~\ref{sec:results_sds}.}

 \section{Image-to-Image Translation}
 \label{section:im2im}

With our Delta Denoising Score, we can apply a direct optimization over the image pixel space, i.e. optimizing for $z = \theta$ as illustrated in Figure~\ref{fig:teaser}.
However, optimizing an image for each editing operation presents several drawbacks. Firstly, it necessitates captions for both the input and the desired edited image. Secondly, the results obtained on real images are inferior to those obtained from synthetic images. Lastly, the time required for inference is long ($\sim$20 seconds per edit). To overcome these limitations, we introduce a novel unsupervised training pipeline for text-driven image-to-image translation based on our proposed \om{}.

\begin{figure}[t]
\footnotesize
\centering
    \begin{overpic}[width=1\columnwidth,tics=10, trim=0mm 0 0mm 0,clip]{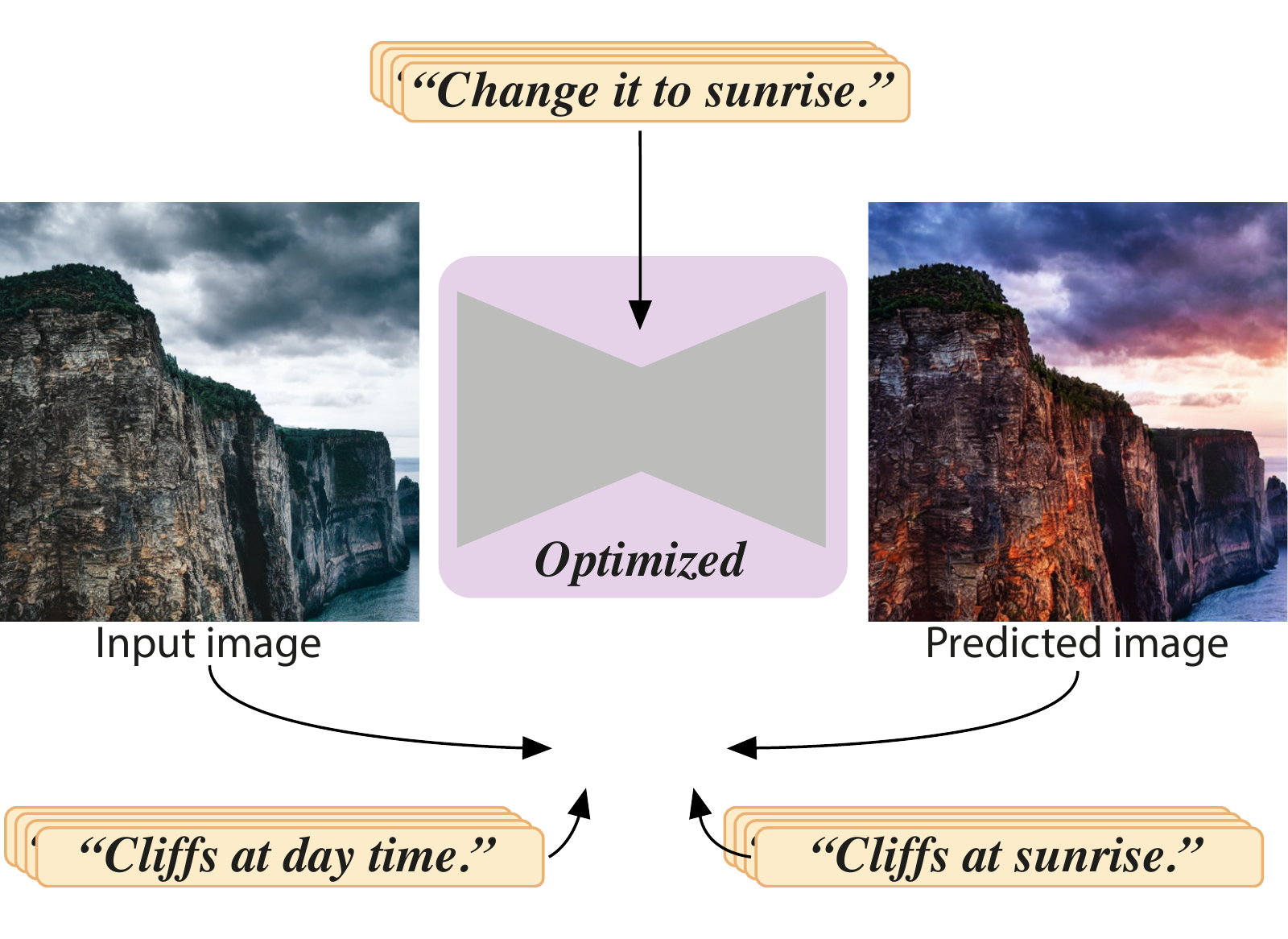}
    \put(15,58){$\latentref$}
    \put(83,58){$\latent$}
    \put(43.5,13){$\nabla_\theta \mathcal{L}_\textrm{DDS}$}
    \put(49,39.5){$\theta$}
    \end{overpic}
\caption{\bf{Unsupervised training for multi task image-to-image translation network.} {\it Given an input image $\latentref$ (left) and a sampled task embedding (top), our network is trained using the Delta Denoising Score (DDS) and corresponding text embeddings (bottom) that describe the input image and the desired edited image result $\latent$. During inference, our network can then translate arbitrary real images based on the specified task, within a single feedforward pass.}}
\label{fig:i2i_diagram}
\end{figure}
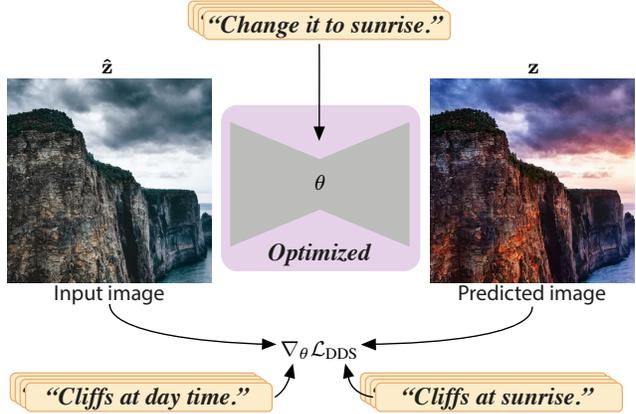

 \paragraph{Unsupervised training with \om}
Using DDS, we introduce an unsupervised training framework for a neural network that learns to translate images based on a caption that describes a \emph{known source distribution} and another caption that describes an \emph{unknown target distribution}.
Given a dataset of source images $\{\latentref_i\}$, source caption $\textref$ and a target caption $\textcond$, our goal is to learn a mapping $\latent = g_{\theta}(\latentref)$ such that $\latent$ has high fidelity to both: the input image $\latentref$ and to the target caption $\textcond$.
As illustrated in Figure~\ref{fig:i2i_diagram}, on the bottom, we utilize the DDS formulation in \eqref{eq:dds} to optimize our network.

Naturally, we can extend the network capabilities to be task conditioned. Under those settings, the network learns a finite set of $M$ image-translation tasks that are defined by multiple target captions $\{\textcond_i\}_{j=1}^{M}$ and corresponding learned task embeddings $\{\task_j\}_{j=1}^{M}$, see Figure~\ref{fig:i2i_diagram}. At each optimization iteration, we sample a source image $\latentref_i$ with its source caption $\textref_i$, a task embedding $\task_j$ with the corresponding target caption $\textcond_j$. Then the network is optimized by the DDS \ref{eq:dds} where $\latent = g_{\theta}(\latentref_i | \task_j)$.

To maintain the fidelity to the input image, we add a weighted identity regularization term:
$$ \mathcal{L}_{\text{ID}} = \lambda_{id}(t) ||g_{\theta}(\latentref_i | \task_j) -  \latentref_i||_{2}^{2},$$
where the weight $\lambda_{\text{ID}}(t)$ is a function of the training iteration $t$, such that at the beginning of the training, we inject prior knowledge on the desired output, and gradually reduce it during training with a cosine decay.

\begin{figure}[t]
    \centering
    \includegraphics[width=\columnwidth]{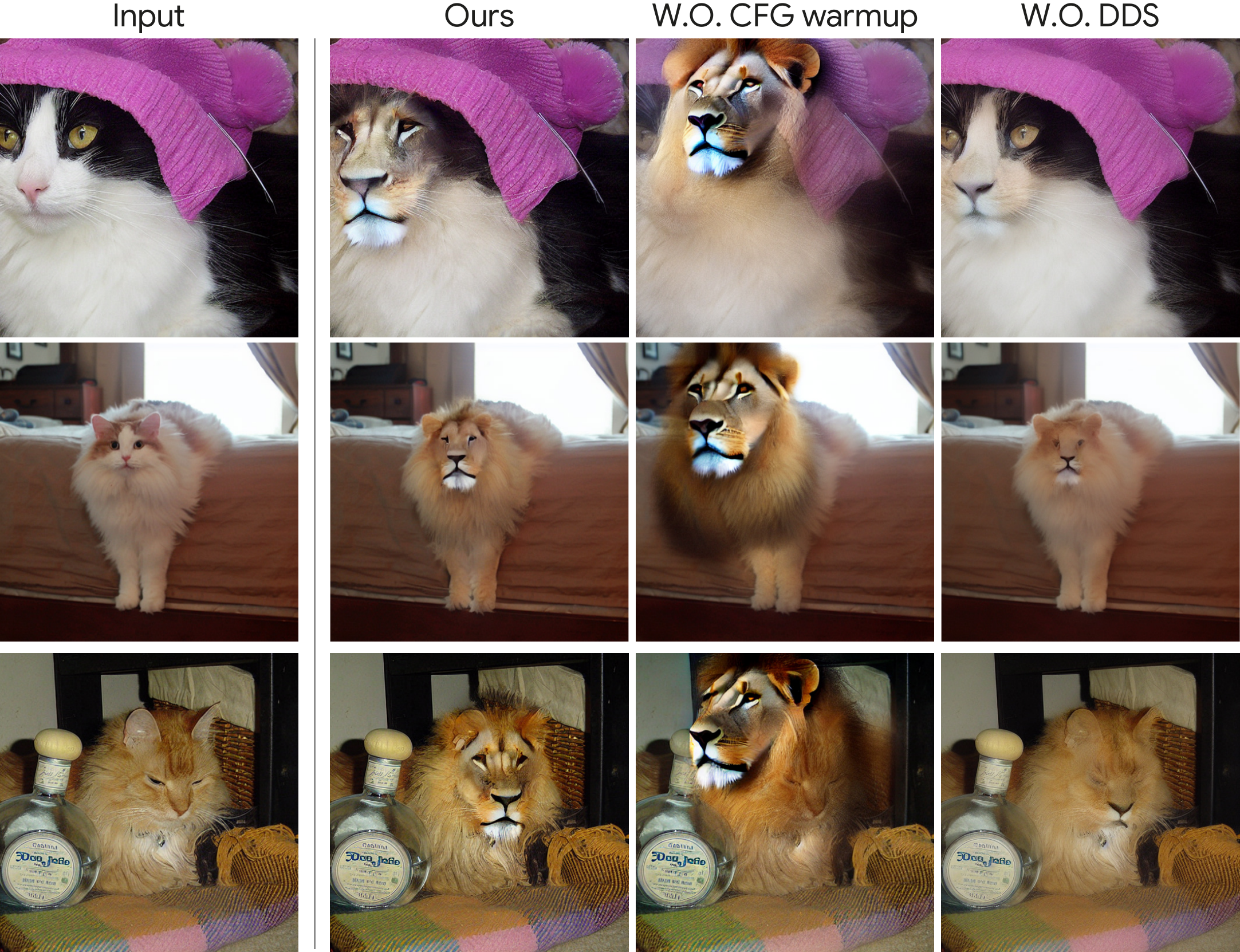}
    \caption{{\bf Ablation study.} \it{We train a cat-to-lion image translation network under various settings. The first and second columns show the input and output results of our full method, respectively. The third row shows the results when training without CFG warmup, and the last column shows the results when training with SDS instead of DDS.}}
    \vspace{-10pt}
\label{fig:ablation}
\end{figure}

\paragraph{DDS with CFG warmup}
During the training of the aforementioned network, we experienced a familiar \textit{mode collapse} phenomena associated with the training of generative adversarial network (GAN)~\cite{goodfellow2014generative}, where the network optimizations led to a local minima.
In our case, the network has learned to produce a fixed object, in a fixed location within the input image, as demonstrated in Figure~\ref{fig:ablation}, where the same type of lion appears in the same pose and locations in all the outputs without respecting the input image.
The reason for the mode collapse in our case can be explained thorough the analogy to GANs.
The discriminator output score that discriminates between real and fake images can be replaced by the delta denoising score. At a local minimum point, our network succeeded to \textit{fool} the DDS such that the output has high fidelity to $\textcond$ at the fixed region and high fidelity to $\latentref$ elsewhere.

To address this issue we have found that implementing a warmup scheduler for the classifier free guidance parameter $\cfg$, utilized in the estimation of the DDS gradient can be effective.
As we have demonstrated earlier, adopting a low value for $\cfg$ during zero-shot optimization is associated with a notably slow convergence rate. Conversely, high values push the image aggressively towards $\textcond$ and lead the training to mode collapse. By gradually increasing the guidance scale, the network gradually learns to make larger changes to the input image with respect to the translation task and avoids local minima.

\section{Evaluations and Experiments}
\label{sec:exp}

In this section we evaluate our observation regarding the SDS and DDS scores, compare our approach to other state-of-the-art zero-shot editing methods and conduct an ablation study to show the effectiveness of different choices in our system.

\paragraph{SDS evaluation}
\label{sec:results_sds}

We measure the expected SDS norm as a function of the timestamp $t$ for \ka{matched and unmatched} image-text pairs. The \ka{matched} pairs obtained by generating images using Stable Diffusion \cite{rombach2021highresolution} with subset of $100$ captions from COCO validation dataset \cite{caesar2018coco}. Then, for each image $\latent$, caption $\textcond$ and timestep $t$ we estimate the value $\mathbb{E}_{\epsilon \sim \mathcal{N}(0, \mathbf{I})} || \nabla_{\latent} \mathcal{L}_{\textrm{SDS}}(\latent, \textcond)||_{2}$ by averaging the result of $200$ measurements and report the average value of the $100$ estimations. To provide a reference, we also perform the experiment on $100$ \ka{unmatched} image-text pairs obtained by permuting the captions of the \ka{matched} set. 
The results are shown in Figure~\ref{fig:sds_graph} (left). As can be seen, SDS exhibits non-negligible high gradients for \ka{matched} pairs. In addition, the gap between \ka{matched} and \ka{unmatched} pairs supports our observation in Section~\ref{section:method} that there is an inherent noise direction $\delta_\textrm{bias}$ in the SDS gradient.     

\paragraph{DDS evaluation}
Next, we evaluate our estimation that for a \ka{matched} pair of similar images with their corresponding text, the SDS noise directions, $\delta_\textrm{bias}$ and $\hat{\delta}_\textrm{bias}$,  are correlated. For this experiment we use a subset of 10000 synthetic image pairs  $\latent$ and $\latentref$ with their corresponding captions $\textcond$ and $\textref$ from InsturctPix2Pix \cite{brooks2022instructpix2pix} dataset .
For each timestamp, we estimate the cosine similarity between $\nabla_{\latent} \mathcal{L}_{\textrm{SDS}}(\latent, \textcond)$ and $\nabla_{\latentref} \mathcal{L}_{\textrm{SDS}}(\latentref, \textref)$ and report the average result across all pairs. Here again, we applied the same experiment to \ka{unmatched} pairs for reference. Note that the caption for each SDS estimation remained aligned to its image.
The results are summarized in Figure~\ref{fig:sds_graph}, on right. As can be seen, the \ka{matched} pairs are strongly correlated which supports our assumption that an estimation for $\delta_\textrm{bias}$ of reference image and text can be used the eliminate the same term from similar pair.

\begin{figure}
    \includegraphics[width=\columnwidth]{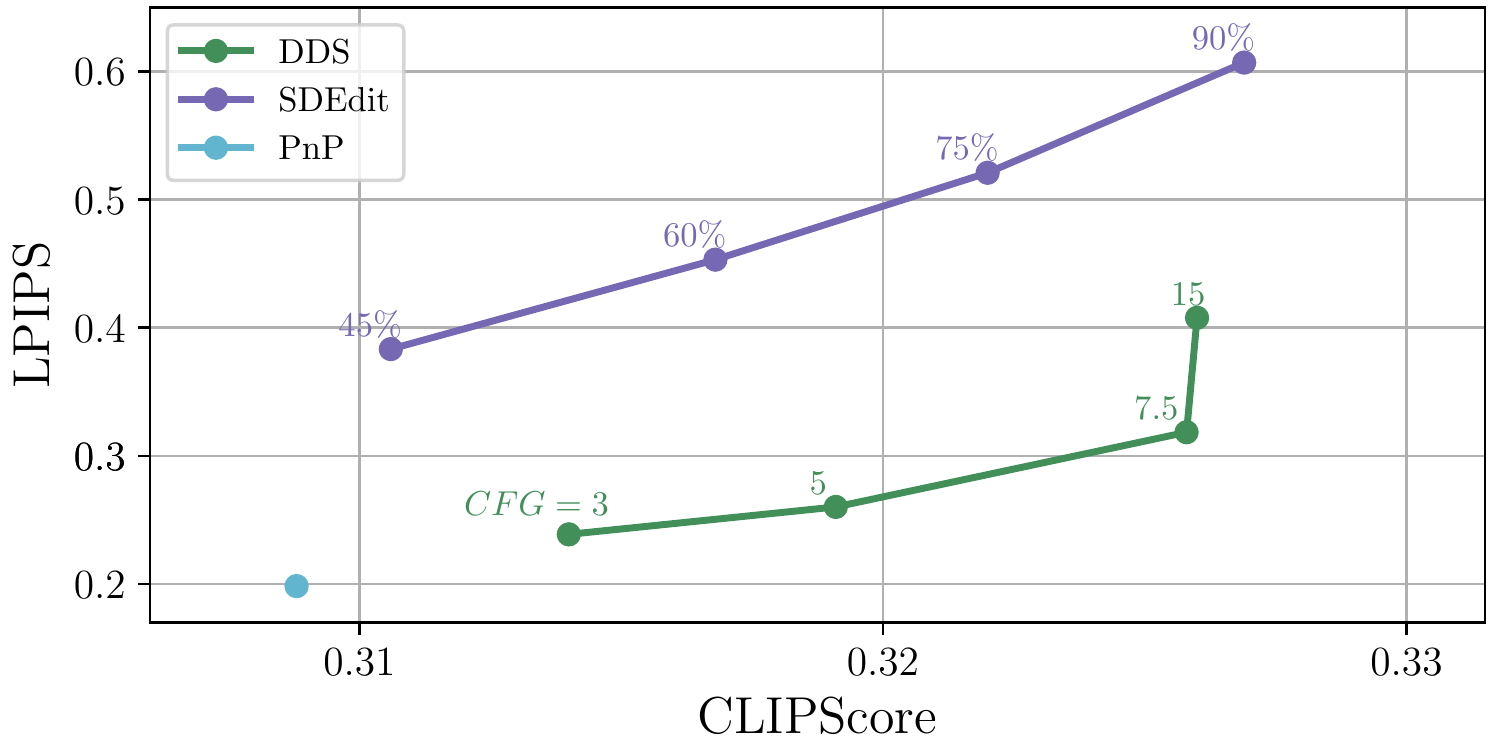}
    \vspace{-3pt}
    \caption{ \bf{Zero shot image editing quantitative comparison.} \it{Using our DDS optimization technique, we tested various CFG values on a dataset of 1000 images and prompts from the InstructPix2Pix training set~\cite{brooks2022instructpix2pix}, and compare our approach to SDEdit~\cite{meng2021sdedit} with different numbers of forward diffusion (noise addition) steps and Plug-and-Play (PnP)~\cite{tumanyan2022plug}. Our outputs have higher fidelity to the source images (low LPIPS scores) while also achieving high fidelity to the edits described in the text prompts (high CLIP scores).}}
    
    \label{fig:zeroshot_comparison}
\end{figure}
\begin{figure}[t]
    \includegraphics[width=\columnwidth]{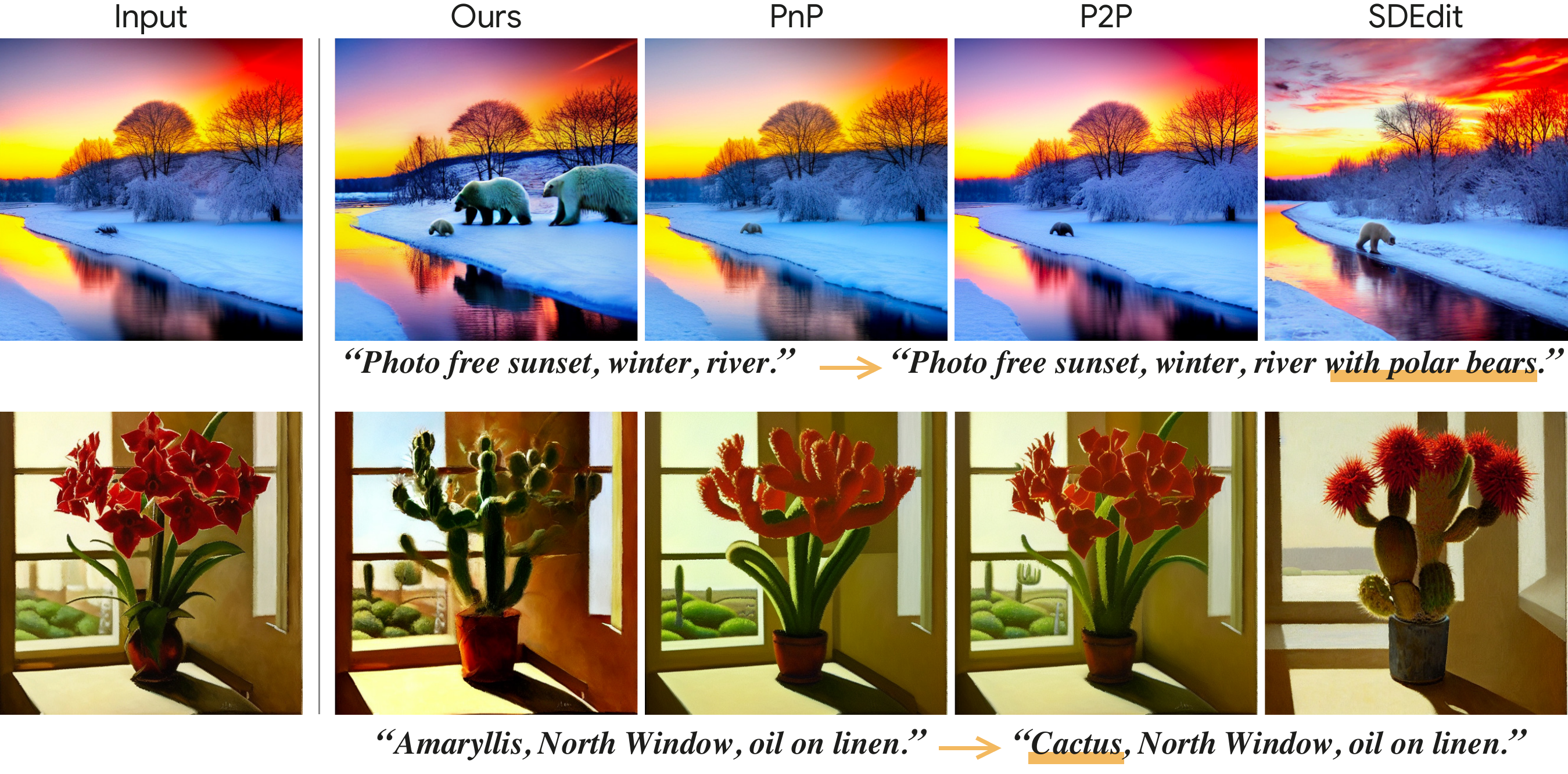}
    \caption{ \bf{Zero shot image editing qualitative comparison.} \it{We compare our approach to SDEdit~\cite{meng2021sdedit} Plug-and-Play (PnP)~\cite{tumanyan2022plug} and prompt-to-prompt~\cite{hertz2022prompt}. Our method showcases its ability to better apply both structural and color changes described in the target text prompt while simultaneously preserving high fidelity to the input image.}}
    \label{fig:zeroshot_qual_comparison}
    \vspace{-6pt}
\end{figure}
\paragraph{Comparison to zero-shot editing methods}
To evaluate our editing capability using a direct DDS optimization over the pixel space of a synthetic generated image, we use a randomly selected subset of $1000$ pairs of source and target prompts from the dataset of InsturctPix2Pix \cite{brooks2022instructpix2pix}. The dataset already includes the paired image results obtained by Prompt-to-Prompt (P2P) \cite{hertz2022prompt}, from which we took only the source images.
For each editing result we measure the text-image correspondence using CLIP score\cite{radford2021learning}. In addition, we evaluate the similarity between the original and the edited images using the LPIPS perceptual distance~\cite{Zhang2018TheUE}. 
We compare our method to additional zeros shot methods: SDEdit~\cite{meng2021sdedit}, and Plug-and-Play (PnP)~\cite{tumanyan2022plug}. It can be seen in Figure~\ref{fig:zeroshot_qual_comparison} that comparing to other methods, our approach demonstrates higher fidelity to the text prompt and to the source image on average. The quantitative results are summarized in Figure~\ref{fig:zeroshot_comparison} where we show the metrics of our method for different numbers of classifier free guidance scale. Notice, that as observed in Figure~\ref{fig:flamingo_cfg}, the improvement to the fidelity to text that obtained by using a large value of CFG is negligible compared to the deterioration in the fidelity to the source image.

\begin{table}
\caption{Quantitative comparison for the multi-task image-to-image translation network. We measure text-image correspondence using CLIP \cite{radford2021learning}. In addition, we evaluate the similarity between the original and the edited images using the LPIPS \cite{Zhang2018TheUE} perceptual distance.}
 \footnotesize
 \centering
 \vspace{0.2cm}
\begin{tabular}{lcc}
\toprule
& CLIP score $\uparrow$ &  LPIPS $\downarrow$   \\
\midrule
PnP & $0.221 \pm 0.036$ & $0.31 \pm 0.075$ \\
InstructPix2Pix & $0.2190 \pm 0.037$ & $0.322 \pm 0.215$ \\
DDS (ours) & $\mathbf{0.225 \pm 0.031}$ & $\mathbf{0.104 \pm 0.061}$   \\
\bottomrule
\end{tabular}
\label{tab:comparion_net}
\end{table}


\begin{figure}
    \includegraphics[width=\columnwidth]{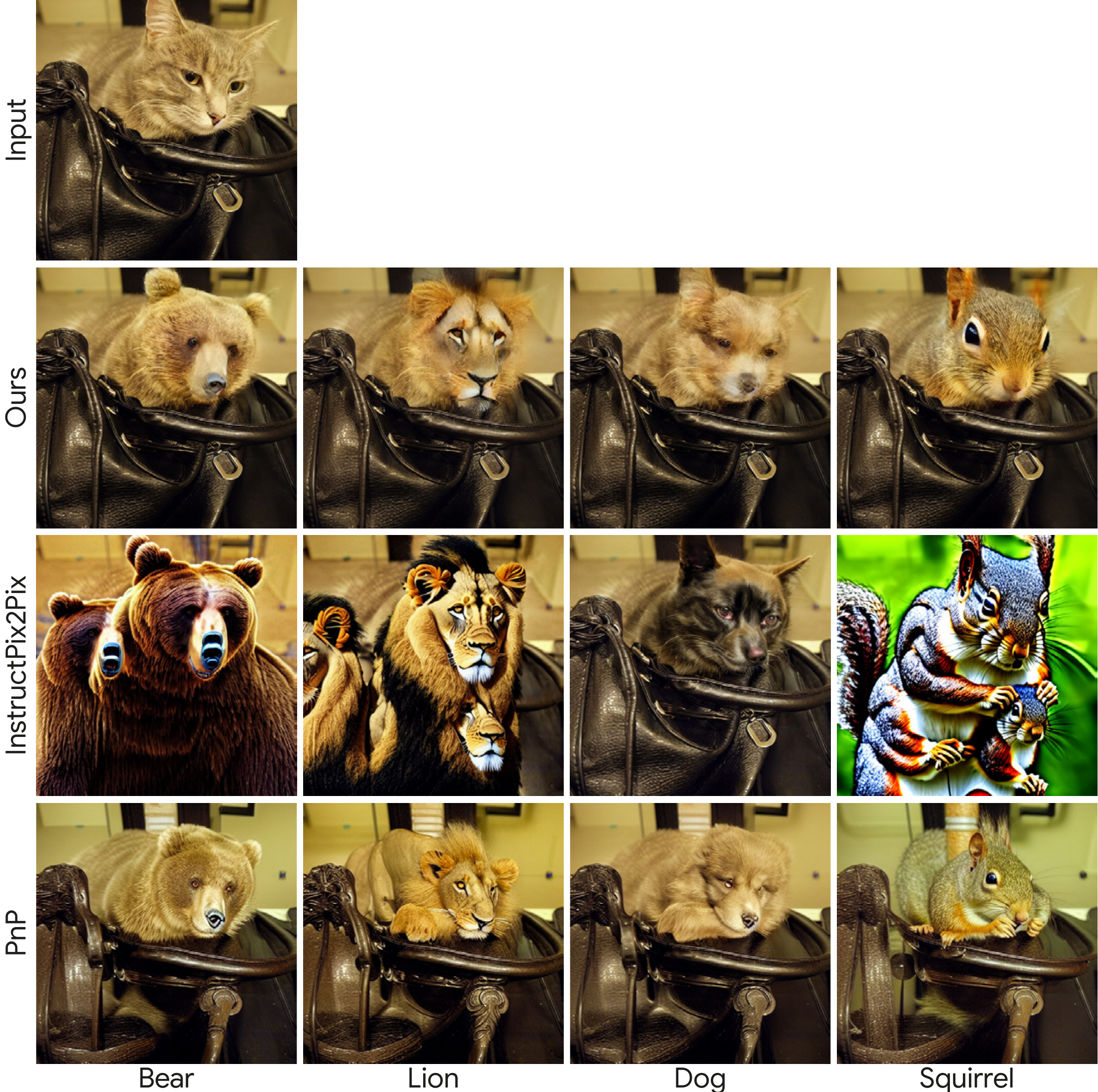}
    \caption{\bf{Image-to-Image translation comparison.} \it{Our multi-task network was trained to translate cats to different animals (lion, dog, squirrel) using DDS. It was trained on synthetic cat photos and evaluated on a subset of real images from the COCO and Imagenet datasets. Our results (second row) better preserve the structure of the cat in the input image as well as its background.}}
    \vspace{-5pt}
    \label{fig:net_compare}
\end{figure}

\paragraph{Image-to-image translation training}

We train different multi-task networks as described in Section~\ref{section:im2im}. For each training instance, we generate a synthetic dataset of $5000$ images using the Stable Diffusion model conditioned on manually written captions (5-20 captions for each dataset). Each training starts from a pre-trained Stable Diffusion model, modified as follows: 
The latent noise inputs are replaced with latents of images from our synthetic dataset.
The text embedding condition is replaced with our learned task embeddings, initialized with text embedding that describes the task embedding. For example, for the task of adding snow in an image, we use the phrase \say{snowing}.
Finally, the timestep of the diffusion process is no longer used since our model inference process contains a single feed-forward. Therefore, we re-use the timestep condition as additional per-task \emph{learned} embedding which is initialized with a positional embedding of $t=0.5$. While the text condition is injected by cross attention, time by adaptive group normalization (ADAGN).
Additional implementation details are provided in the supplementary material. 

\vspace{-6pt}

\paragraph{Image-to-image translation comparison} We evaluate a \emph{Cat-to-Other} network trained to translate images of cats to images of four different animals: a dog, a lion, a bear, and a squirrel.
We tested our network using a collection of $500$ cat images from ILSVRC \cite{ILSVRC15} and COCO \cite{caesar2018coco} validation set; overall, we tested the results of $2000$ image translations.
We use the same LPIPS and CLIP scores to estimate fidelity to the source image and target text that describes the target distribution, for example, \say{A photo of a lion}.
We compare our method to PnP \cite{tumanyan2022plug} and InstructPix2Pix \cite{brooks2022instructpix2pix} which also utilize the generative prior of a pre-trained diffusion model for the task of image-to-image translation. Unlike our method, InstructPix2Pix fine tune a diffusion model using synthetic pairs of images and therefor it is sensitive to quality of the pairing method.

The results are summarized in Table~\ref{tab:comparion_net} and Figure~\ref{fig:net_compare}. As can be seen, our method achieves both: better fidelity to the input image and to desired target domain. Additionally, our method operates via a single feed-forward pass during inference, making it $\times 50$ faster than the other diffusion-based methods that require a full iterative diffusion process in each inference sampling.
A qualitative comparison is shown in Figure~\ref{fig:net_compare}. 
As can be seen, our method better preserves the structure of the cat when translating to other animals. Moreover, our distilled training results in a more robust network that can better distinguish between regions in the image that had to be changed and areas to preserve.

\vspace{-6pt}
\paragraph{Ablation Study}
We evaluate key components of our image-to-image translation network on a single task of cat-to-lion image translation. First, we show our results without the CFG scaling warmup. As shown in Figure~\ref{fig:ablation} (third column), it results in mode collapse where roughly the same lion appears in the same location regardless of the cat in the input.
In addition, we train a network with the \emph{Vanilla} SDS term instead of our DDS while the other components, the $\mathcal{L}_{\text{ID}}$ term and the CFG warmup, remain untouched and prevent the mode collapse. As can be seen (right column in Figure~\ref{fig:ablation}), the quality of the translation to a lion is worse than our full settings. Moreover, the SDS training struggles to preserve high-frequency details in the input image. For example, see the patterns of the purple wool hat in the first row.

\section{Conclusions , Limitations and Future work}

\begin{figure}[t]
    \includegraphics[width=\columnwidth]{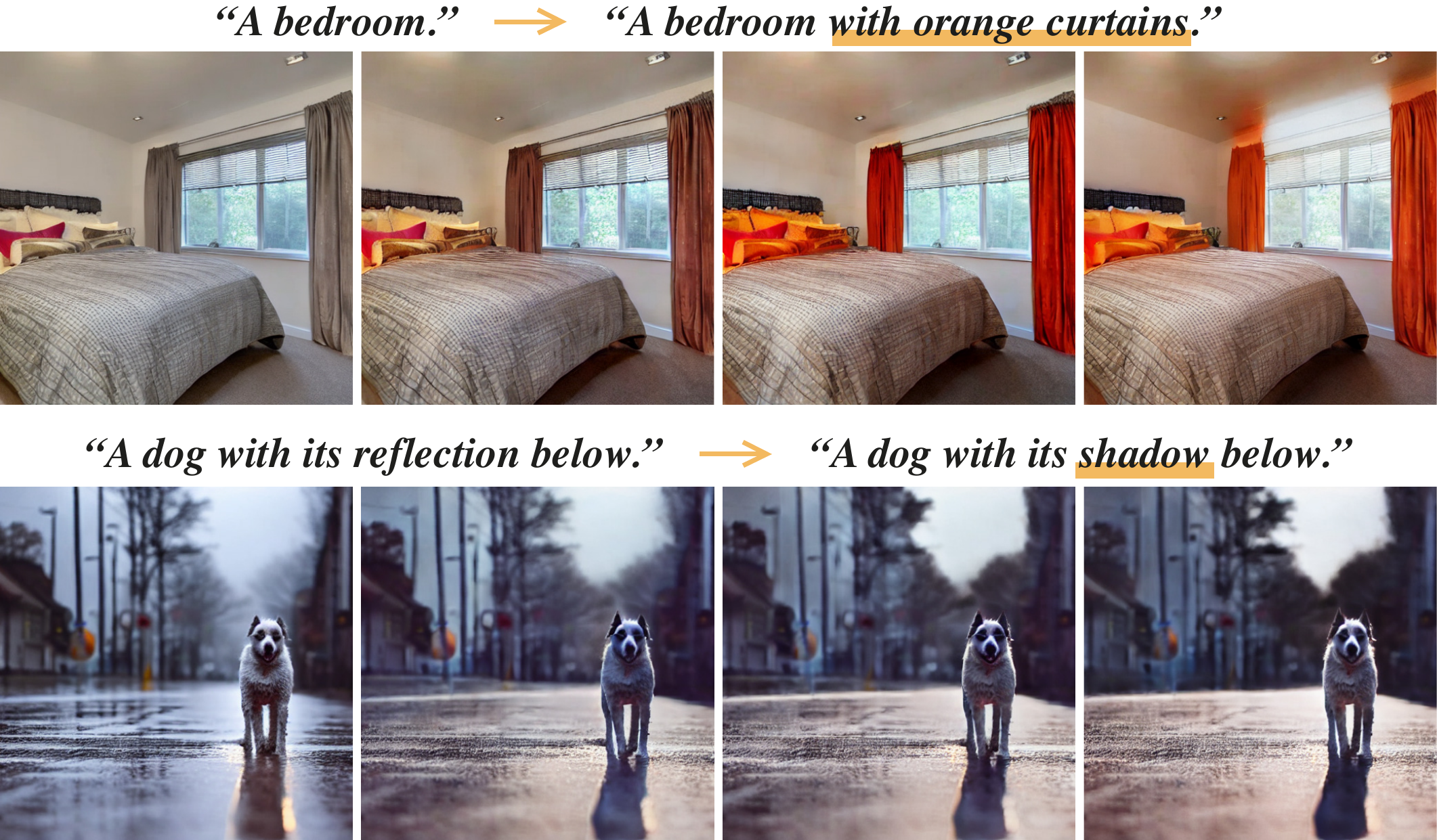}
    \caption{\bf{Limitations.} \it{Biases of the diffusion model or limitations in language understanding affect the DDS optimization. Top: we would like to change the color of the curtains in the bedroom, but the color of the pillows is also changed. Bottom: replacing the dog's reflection with a shadow also causes changes in the lighting, weather and background details.}}
    \vspace{-10pt}
    \label{fig:limietaion}
\end{figure}

We have presented, \ourmethod{}, a new diffusion scoring technique that allows optimizing a given image as means to edit it with respect to a text-prompt. \ourmethod{} uses the SDS score applied to input image to calculate cleaner gradients during the optimization, which leads to a distilled edit. We have also showed an image-to-image translation model trained with our new score. The model is training with no supervision, requires no pairs of images, and thus can be trained on real images. 

Our \ourmethod{} works well in distilling text-driven image-to-image translations. However, there are cases that the results are imperfect due to the inherent limitation of the text-to-image model, in particular its language model. A noticeable problem is the binding of adjectives to the nouns. A typical example is demonstrated in Figure \ref{fig:limietaion} (top), where the orange color is not well bind to the curtains, and thus licks to the entire bedroom. Another example, displayed in the bottom row, where the dog's reflection is replaced with a shadow and caused unwanted changes in the background details.

We also acknowledge that the multi-task model can be better trained and may be further improved by combining multiple experts training \cite{balaji2022eDiff}, which uses multiple network, or utilize subset of paired data and train our network under semi-supervised settings.

The scope of \ourmethod{} is wide, and  its generalization across various editing tasks \cite{brooks2022instructpix2pix} should be explored in the future. Furthermore, we believe that it can be extended to other modalities, such as text-driven 3D shape editing, video editing and motion editing \cite{metzer2022latent, kasten2021layered, tevet2022motionclip}.

The objective of this work is to extract efficient and clean gradients that can facilitate the optimization of an image towards a distilled edit. This, we believe, is an important step towards enhancing our understanding of how to effectively extract and utilize the rich knowledge that is concealed within large-scale generative models.

\section{Acknowledgement}
We thank Ben Poole, Elad Richardson, Ron Mokady and Jason Baldridge for their valuable inputs that helped improve this work.


{\small
\bibliographystyle{ieee_fullname}
\bibliography{egbib}
}

\appendix
\noindent{\Large \textbf{{Appendix}}}

\section{Societal Impact}
Our method introduces an unsupervised editing framework for images. Our framework might be exploited for producing fake content, however, this is a known problem, common to all image editing techniques.
Moreover, our method relies on generative priors of a large text-to-image diffusion models that might contain undesired biases due to the auto-filtered enormous dataset that they were trained on.
Those undesired biases could infiltrate to our editing results thorough our optimization process or our distilled networks.
However, we believe that our DDS optimization technique can help in the future to reveal such undesired biases and editing directions, similarly to the way it operates today to clean noisy undesired components within editing directions.

\section{Implementation Details}

\paragraph{Zero-shot DDS optimization}
Unless specified otherwise, in all of our zero shot experiments, we initialize the latent image $\latent$ by the reference latent image $\latentref$ and apply our DDS optimization for 200 iterations ($\sim 18$ seconds on a single A100 GPU) using SGD with learning rate of $2$ and applying learning rate decay of $0.9$ in intervals of $20$ iteration.

\paragraph{Image-to-image network training}
Our image-to-image networks were trained using a batch size of $2$ for $125000$ iterations ($\sim 23$ hours on a single A100 GPU).
We use the Adam optimizer with a learning rate of $10^{-5}$ and learning rate warmup over $10000$ iterations using a linear scheduler.
For the DDS, we set the classifier-free guidance scale (CFG) to $25$ using CFG  warmup starting from $1$ over $20000$ iterations using a cosine scheduler.
For the identity regularization, we start with $\lambda_\textrm{id}=3$ and cool it down to $0.1$ over $20000$ iterations using a cosine scheduler.

\section{Additional Ablations}

\begin{figure*}
    \includegraphics[width=\textwidth]{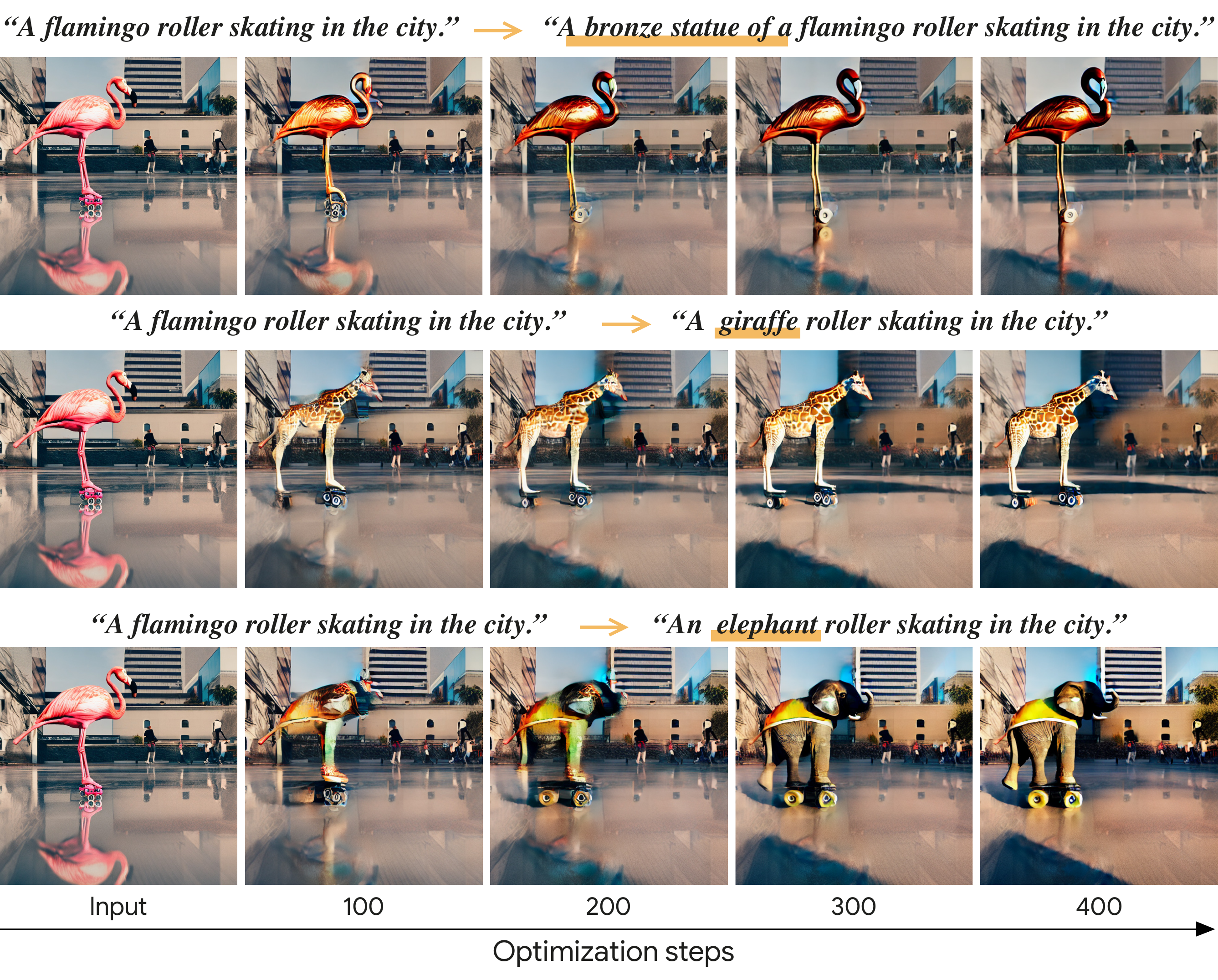}
    \caption{\bf{DDS Optimization convergence.} {\it Some challenging edits, such as changing a flamingo to an elephant, may require more time to converge compared to other edits. In most cases, we have found that $200$ steps are sufficient.}}
    \label{fig:supp_steps}
\end{figure*}

\paragraph{Number of optimization steps}
We set the number of optimization steps to $200$, as we found it sufficient for most desired edits. However, for obtaining large structural changes or color modifications, the number of optimization steps should be increased.
See example in Figure~\ref{fig:supp_steps} where structural modification, of replacing a flamingo to an elephant requires $400$ steps, while modify the flamingo to a bronze sculpture converges faster.
Notice that in some cases, where we seemingly request for a large modification, like replacing the flamingo to a giraffe, $200$  optimization are still sufficient since the overall structure of the giraffe is similar to that of the flamingo.  

\begin{figure*}
\footnotesize
    \centering\begin{overpic}[width=\textwidth,tics=10, trim=0mm 0 0mm 0,clip]{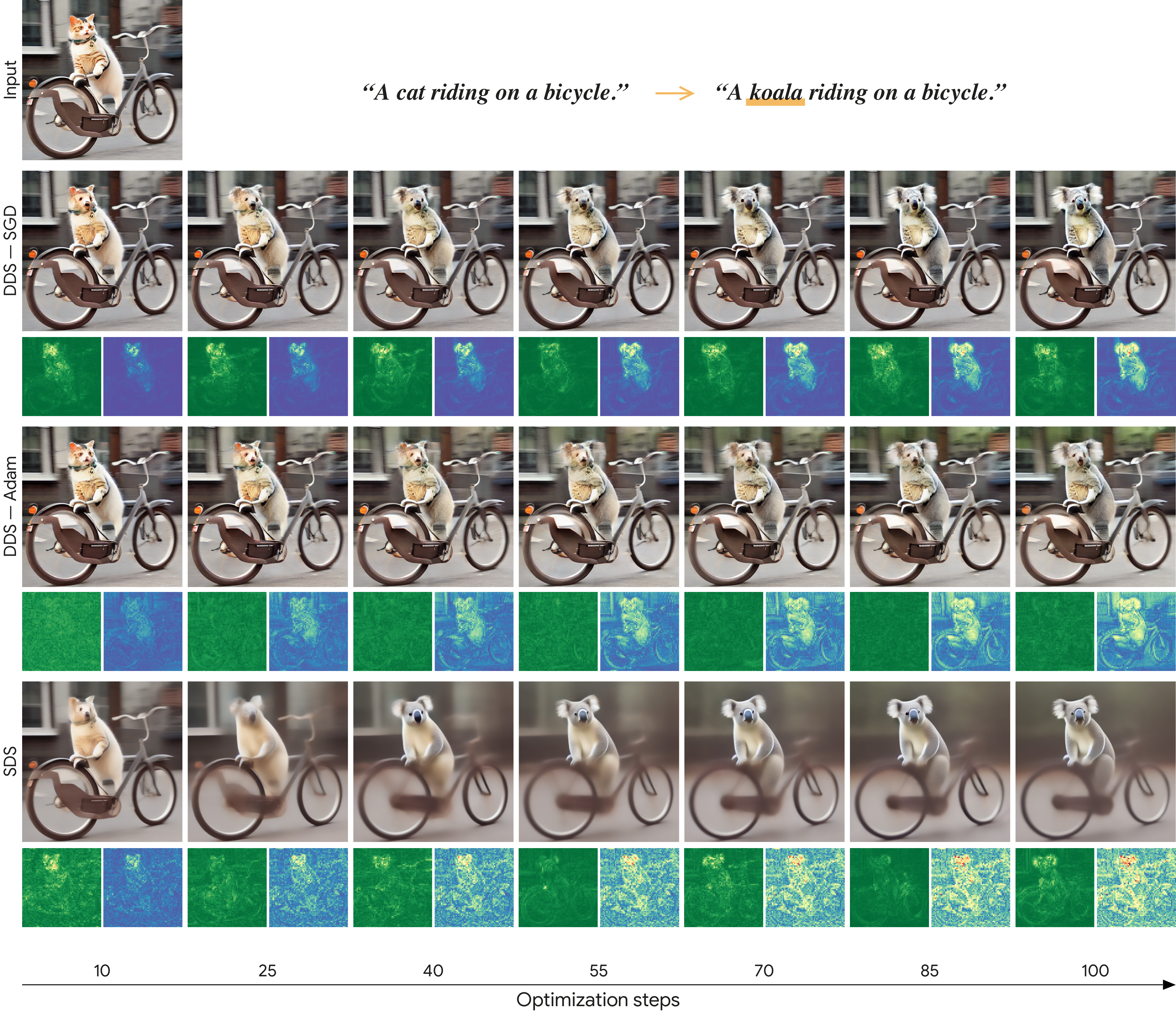}
    \put(4.5,6.5){$\Delta \latent $}
    \put(10,6.5){$\sum\Delta \latent $}
    \end{overpic}
    \caption{\bf{Optimizer selection}. {\it Each row shows a zero-shot editing optimization sequence using different settings.
    The green-to-red \ka{heatmaps} (left map beneath each image) show the norm of the latent pixels update in a specific step. The blue-to-red heatmaps (right map) show the accumulated difference norm with respect to input image (top).
    }}
    \label{fig:supp_grad_vis}
\end{figure*}

\paragraph{Optimizer Selection}
We tested our DDS optimization using both SGD and Adam \cite{kingma2020method} optimizer and found that using \emph{vanilla} SGD leads to higher quality results.
Figure~\ref{fig:supp_grad_vis} compares the two alternatives (top rows) with respect to the baseline SDS with SGD optimization (bottom row).
We can clearly see that both DDS optimization achieve higher quality results compared to the baseline. However, by looking at the accumulated difference of the image during the optimization (right saliency map beneath each image), we can see that the Adam based optimization results with more changes and artifacts that are not related to editing prompt.
The reason for the change in the quality can be explained through the adaptive nature of Adam. For simplicity, consider the simpler update rule of Adagrad \cite{JMLR:v12:duchi11a}:
$$
\theta_t \leftarrow \theta_{t -1} - \gamma\dfrac{g_t}{\sqrt{\sum_{i=1}^{t} g^{2}_{t}}},
$$
where $t$ enumerates the optimization steps, $g_t$ is the gradient of $\theta$ calculated at step $t$ and $\gamma$ is the optimization learning rate. We can see that
the normalization to the gradients may magnify outlier gradients or reduce the weight of \emph{good} gradients. 
Figure~\ref{fig:supp_grad_vis} visualize such update to the pixels for different steps across the optimization (left saliency map beneath each image). It can be seen that the Adam based optimization (middle row) leads to a uniform update across the pixels compared to SGD (top row), where most of the energy is located at the pixels that are relevant to the edit.

\begin{figure*}
    \centering
    \begin{overpic}[width=\textwidth,tics=10, trim=0mm 0 0mm 0,clip]{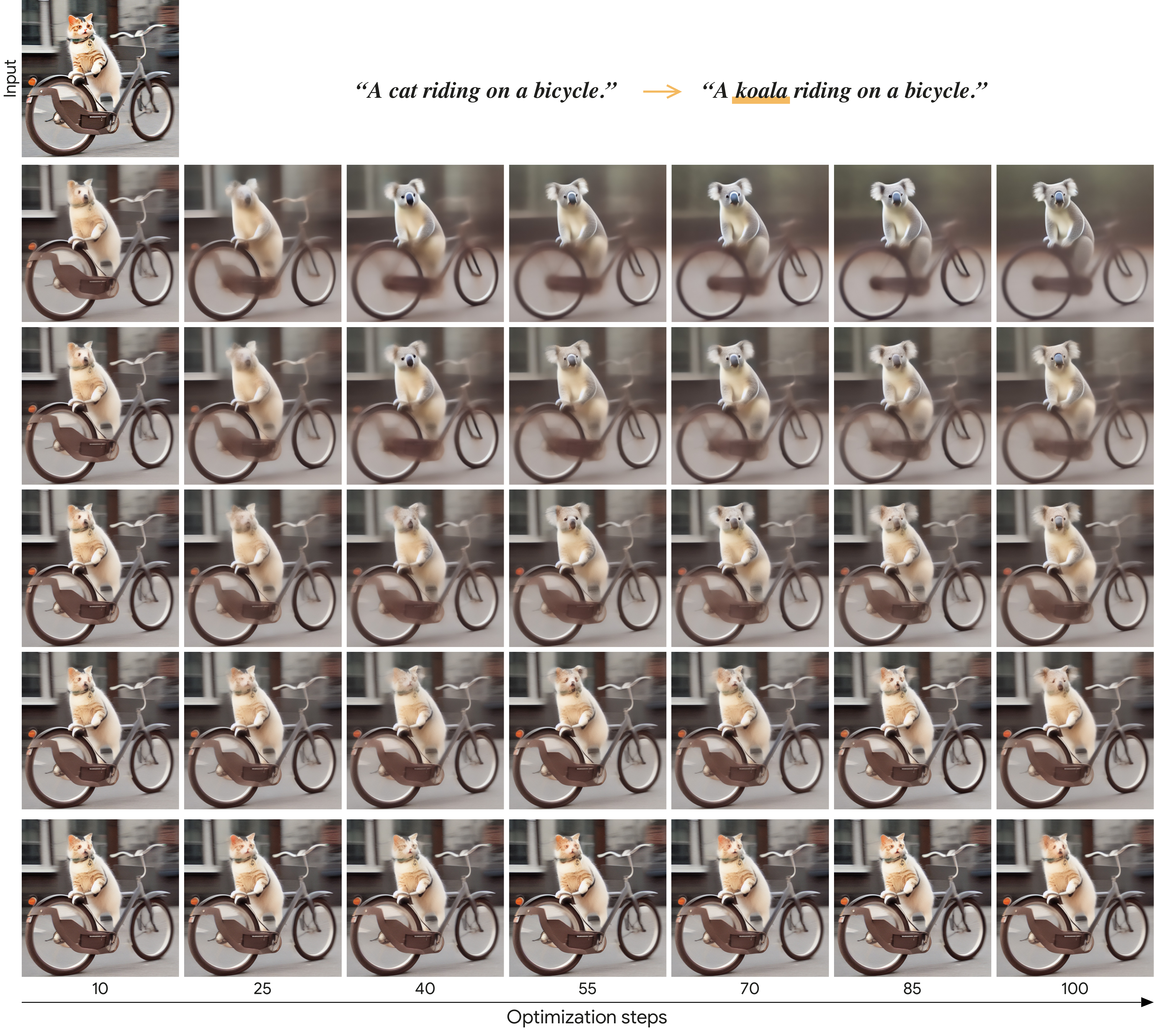}
    \put(0,65.5){\rotatebox{90}{$\lambda_\textrm{id}=0$}}
    \put(0,54.5){\rotatebox{90}{$1$}}
    \put(0,40.5){\rotatebox{90}{$2$}}
    \put(0,26.5){\rotatebox{90}{$5$}}
    \put(0,11.5){\rotatebox{90}{$10$}}
    \end{overpic}
    \caption{\bf{Regularized SDS optimization as a baseline}. {\it We combined a \emph{vanilla} SDS optimization with a weighted $\mathcal{L}_2$ loss between the optimized image and the input image (top).
    From top row to bottom: we use a larger weight ($\lambda_\textrm{id}$) for the $\mathcal{L}_2$ loss. As can be seen, by modifying $\lambda_\textrm{id}$ we can trade off between fidelity to the text prompt and fidelity to the input image.}}
    \label{fig:supp_sds_reg}
\end{figure*}

\paragraph{Regularized SDS optimization}
An alternative to our DDS approach is to use the SDS with additional regularization that prevents large changes in the edited image $\latent$ with respect to the input $\latentref$. The simplest choice is to \st{add} regularise the optimization with a weighted $\mathcal{L}_2$ loss between $\latent$ and  $\latentref$. 
In this setting, the gradient of $\latent$ is given by:

$$\nabla_{\latent} = \nabla_{z}\textrm{SDS} + \lambda_{id} \left(\latent - \latentref\right),$$
Where $\lambda_{id}$ is the weight for the regularisation $\mathcal{L}_2$ loss.
Figure~{\ref{fig:supp_sds_reg}} shows the results of regularized SDS optimizations using increasing weight for the regularization term. As expected, increasing the value of $\lambda_{id}$ harms the fidelity to the edit prompt, while the blurriness side-effect of SDS cannot be avoided.

\section{Additional Results}
\begin{figure*}
\centering
    \includegraphics[width=\textwidth]{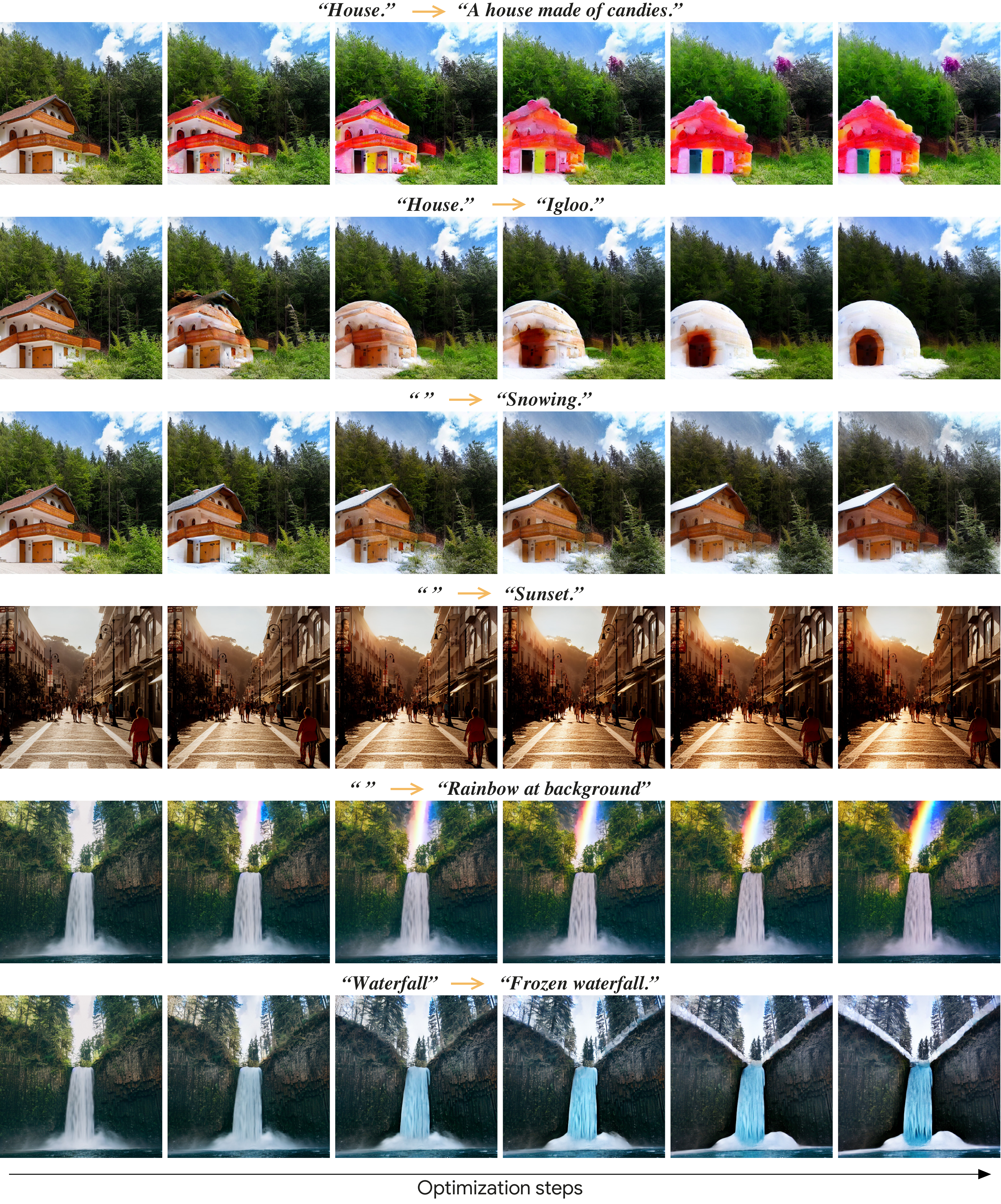}
        \caption{\bf{Zero-shot image editing using DDS optimization \ka{on real images}.} {\it Using DDS, we can apply a variety of edits over \ka{real} images using simple \ka{input prompt descriptions (like ``house", or even an empty prompt)}. The editing operations are mask free and may contain global descriptions, for example, changing the lighting in the image,  or local descriptions, such as changing a house to an igloo.} }
    \label{fig:supp_zero1}
\end{figure*}

\begin{figure*}
\centering
    \includegraphics[width=\textwidth]{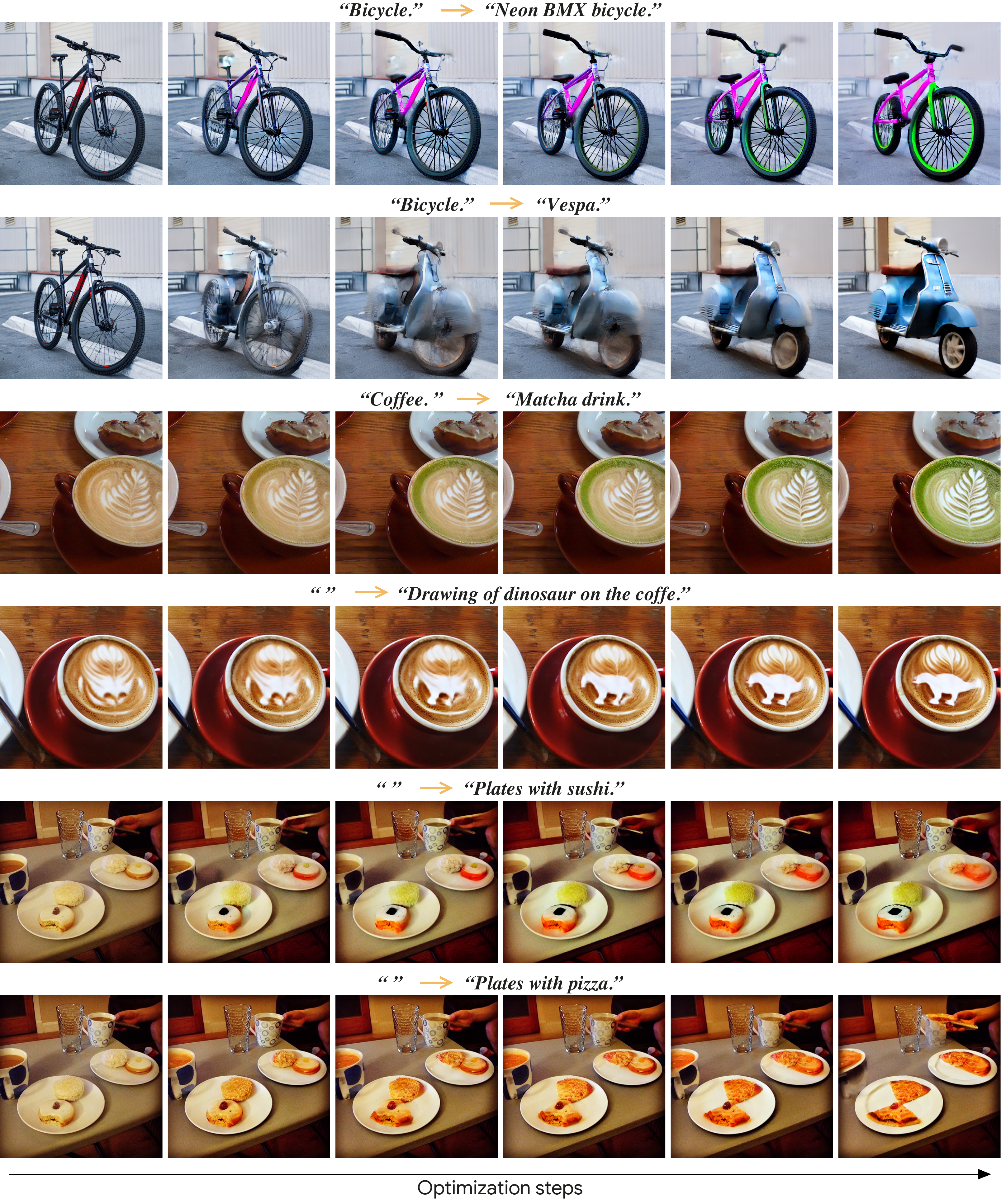}
    \vspace{-10pt}
    \caption{\bf{Zero-shot image editing using DDS optimization \ka{on real images}.} {\it Using DDS, we can apply a variety of edits over images using simple \ka{input prompt descriptions (like ``Coffee", or even an empty prompt)}. The editing operations are mask free and may contain structural changes, for example, changing a bicycle to a Vespa, or stylistic changes, such as changing coffee to a matcha drink.}}
    \label{fig:supp_zero2}
\end{figure*}

\paragraph{Zero-shot image editing results \ka{on real images}}
Additional image editing results using DDS optimization are shown in Figures~\ref{fig:supp_zero1} and \ref{fig:supp_zero2}. All results applied on real images from COCO, and Unsplash datasets \cite{caesar2018coco, unsplash_2022}. Notice that our method works with simple input prompts describing the edit we want to apply to the image. We use a reference text  $\textref$ only in cases where we want to apply the edit over a specific object. Otherwise, we set $\textref$ to the embedding of the null text. 

\begin{figure*}
\centering
    \includegraphics[width=\textwidth]{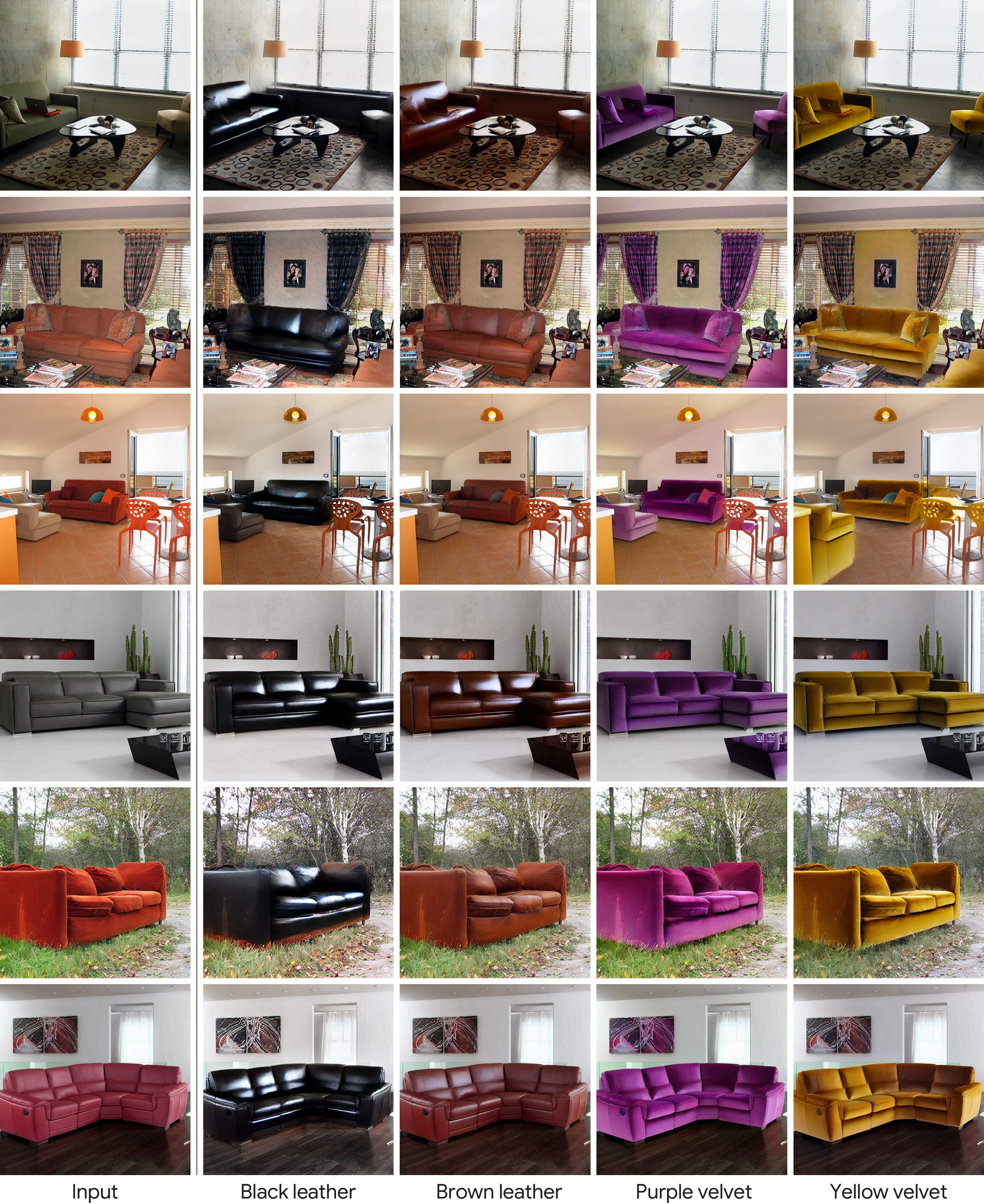}
    \caption{\bf{Unsupervised multi-task image-to-image translation-- sofas network results.} {\it The network was trained to change the color and material of the sofa in the input image.
    The network was trained on synthetic images of living rooms and tested on real living rooms images from ILSVRC \cite{ILSVRC15} and COCO \cite{caesar2018coco} validation sets.}}
    \label{fig:supp_sofas}
\end{figure*}

\begin{figure*}
\centering
    \includegraphics[width=\textwidth]{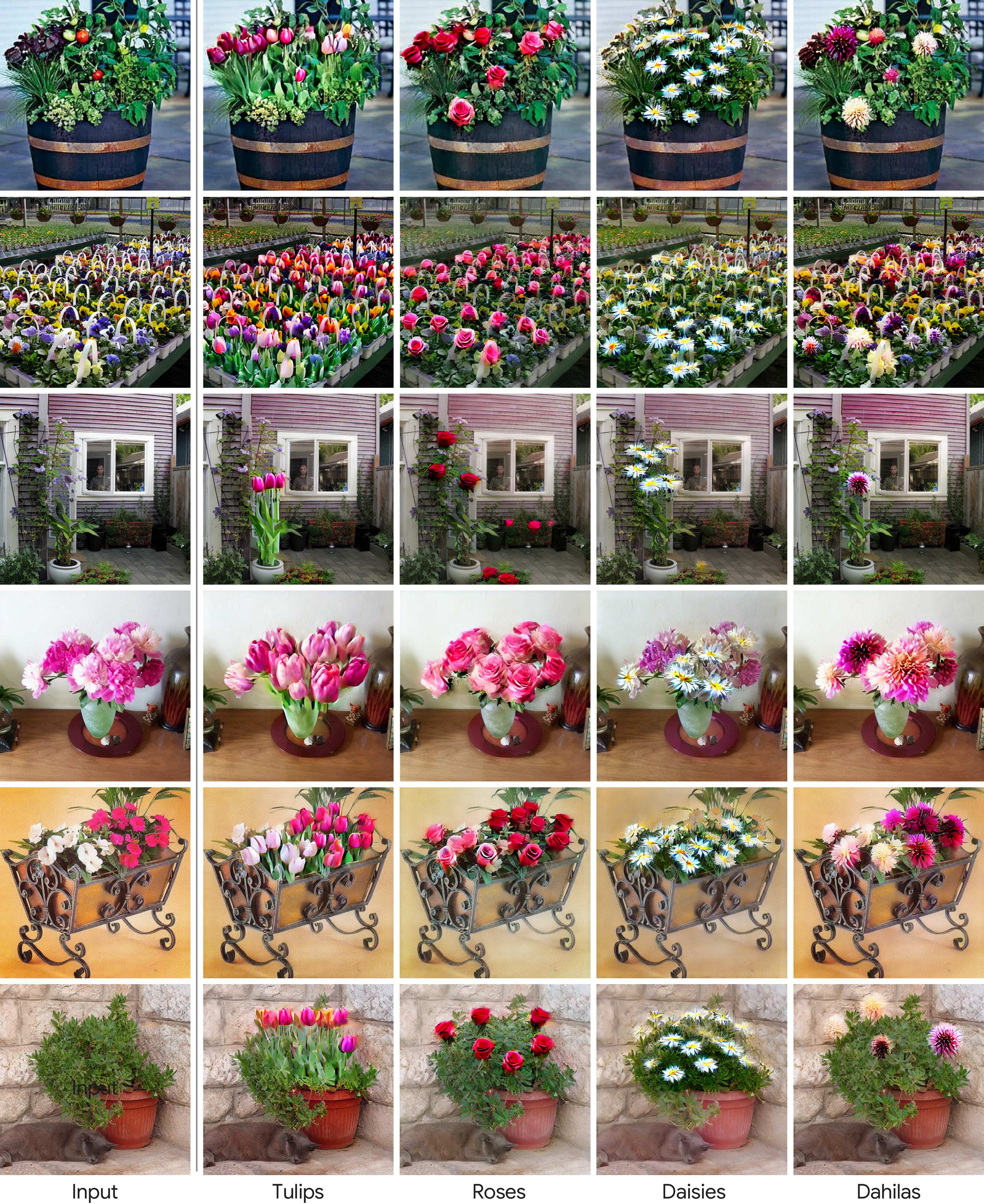}
    \caption{\bf{Unsupervised multi-task image-to-image translation-- flowers network results.} {\it The network was trained to change to add different flowers potted plant in the input image.
    The network was trained on synthetic images and tested on real images of flowerpots \cite{ILSVRC15} and potted plant \cite{caesar2018coco}.}}
    \label{fig:supp_flowers}
\end{figure*}

\paragraph{Image-to-image networks results}
Additional results are shown in Figures~\ref{fig:supp_sofas} and \ref{fig:supp_flowers}.
The first network was trained to change the material of a sofa in an input image. The network was trained on a synthetic dataset containing $5000$ images of living rooms.
An additional network was trained to synthesize different flowers in images of potted plants. The network was trained on a synthetic dataset containing $5000$ of potted plants in living rooms, kitchens, in gardens, and in city streets.
Finlay, we train different networks to modify a person in an input image to other characters. Synthesizing images of persons using Stable Diffusion usually results in poor results. Therefore we train our \emph{character} networks over FFHQ in-the-wild dataset (unaligned) \cite{karras2019style}. During training, we set a single pair of fixed prompts for the DDS optimization, \say{A photo of a person.}  for the embedding $\textref$, and for the target embedding $\textcond,$ we replace the word \say{person} with one of the characters: a Claymation character, A sculpture, a 3D Pixar character and to a zombie. All results are shown over images from ILSVRC \cite{ILSVRC15} and COCO \cite{caesar2018coco} datasets.


\newpage


\end{document}